\documentclass[11pt]{article}

\usepackage[final]{acl}

\usepackage{times}
\usepackage{latexsym}

\usepackage[T1]{fontenc}

\usepackage[utf8]{inputenc}

\usepackage{microtype}

\usepackage{inconsolata}

\usepackage{graphicx}

\usepackage{algorithm}
\usepackage{algorithmic}
\usepackage{amsmath}
\usepackage{amssymb}
\usepackage{kotex}
\usepackage{xcolor}
\usepackage{booktabs}
\usepackage{multirow}
\usepackage{arydshln}
\usepackage{xspace}
\usepackage{arydshln}
\usepackage[table]{xcolor}
\usepackage{titletoc}

\newcommand{\MODELNAME}{\textsc{Blanc}\xspace}

\title{BLANC: Adaptive Temporal Scheduling for Sign Language Production}

\author{First Author \\
  Affiliation / Address line 1 \\
  Affiliation / Address line 2 \\
  Affiliation / Address line 3 \\
  \texttt{email@domain} \\\And
  Second Author \\
  Affiliation / Address line 1 \\
  Affiliation / Address line 2 \\
  Affiliation / Address line 3 \\
  \texttt{email@domain} \\}

\author{
 \textbf{Youngmin Kim\textsuperscript{1}},
 \textbf{Seoyeong Hwang\textsuperscript{2}},
 \textbf{Sujung Hong\textsuperscript{1}},
 \textbf{Seong Jae Hwang\textsuperscript{1}}
\\
\\
 \textsuperscript{1}Yonsei University \quad
 \textsuperscript{2}Ewha Womans University
\\
 \small{
   \texttt{\{winston1214, sujung0914, seongjae\}@yonsei.ac.kr\textsuperscript{1} \quad se00@ewha.ac.kr\textsuperscript{2}}
 }
}

\begin{document}
\maketitle
\begin{abstract}
Sign language production (SLP) aims to synthesize expressive signing motion from written text. Modeling signing motion presents two key temporal challenges. First, fine-grained hand articulations demand higher temporal resolution than relatively slower body movements. Second, signing events unfold over variable durations, calling for adaptive temporal scheduling during generation. To address these challenges, we first introduce a Multi-Rate Vector Quantizer (MRVQ), an articulation-decoupled motion tokenizer that represents the body and both hands as separate discrete streams at distinct temporal resolutions. Building on this representation, we propose \textsc{Blanc} (\textbf{BL}ock-wise \textbf{A}daptive diffusio\textbf{N} with Articulatory \textbf{C}lock), a gloss-free SLP framework that generates variable-duration motion blocks causally while denoising tokens in parallel within each block. At its core, a Latent Articulatory Clock (LAC) learns temporal structure from multi-scale motion-derived partitions to predict block durations and sequence termination during generation. We further introduce inference-time body-conditioned hand refinement to improve coordination across articulators. Extensive experiments on PHOENIX-2014T and CSL-Daily demonstrate that \textsc{Blanc} achieves state-of-the-art performance in sign language production.
\end{abstract}

\section{Introduction}

Sign language is a primary means of communication for Deaf and Hard-of-Hearing (DHH) communities~\cite{padden1988deaf}, making automatic sign language technologies important for improving access to information and communication~\cite{yin2021including}. While substantial progress has been made in sign language understanding and translation~\cite{phoenix, llm_slt,uni_sign}, generating expressive sign motion directly from spoken-language text remains comparatively challenging. Sign language production (SLP)~\cite{progtrans,signingatscale,soke} addresses this problem by synthesizing articulated signing motion from text. The difficulty lies in realizing linguistic content as continuous motion: the model must preserve the intended meaning while inferring the detailed poses, transitions, and timing that are not explicitly specified by the input text.

\begin{figure}[t]
    \centering
    \includegraphics[width=\linewidth]{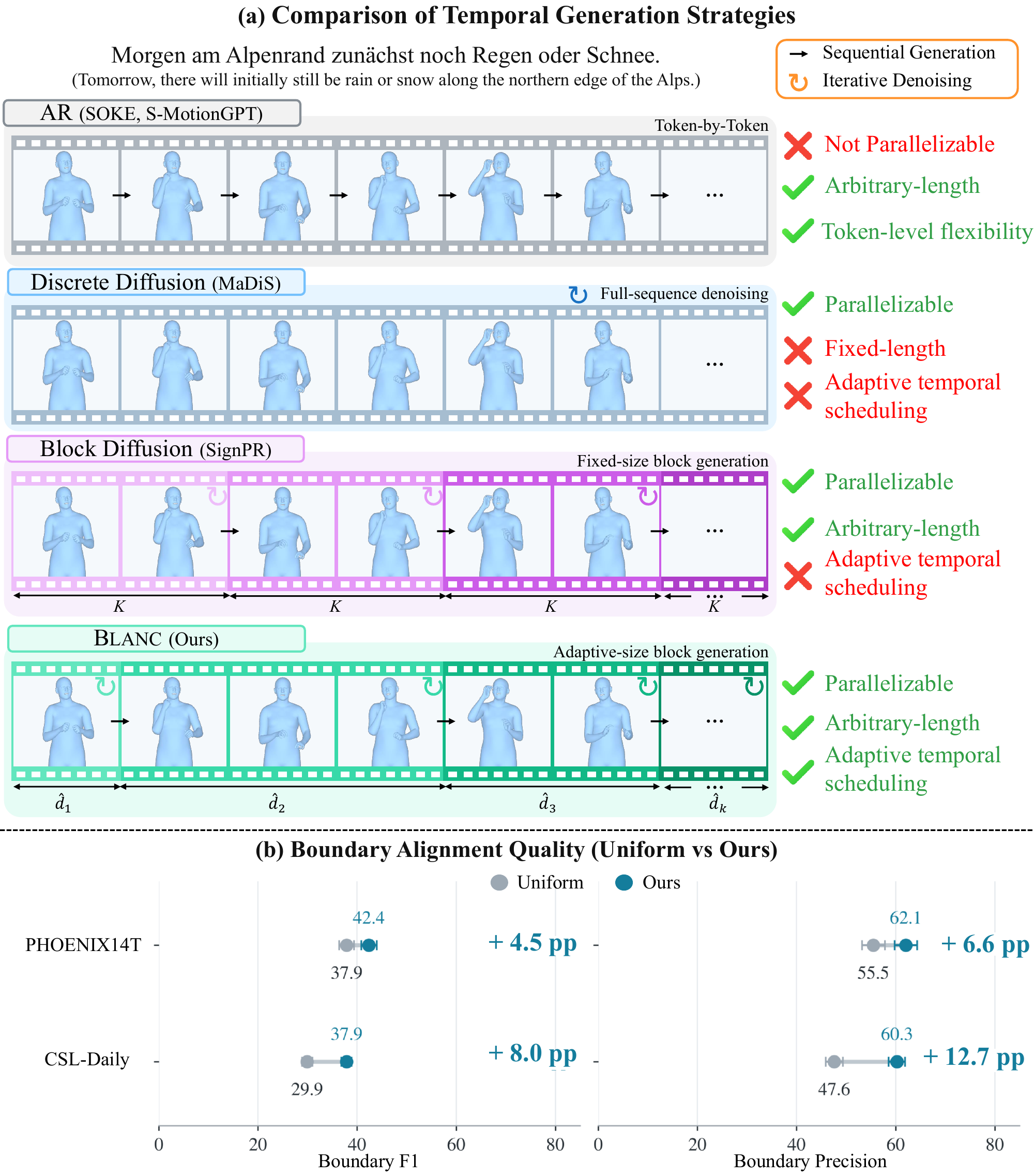}
    \caption{Temporal generation strategies and boundary alignment. (a) Comparison of temporal generation strategies with different generation units and scheduling schemes. (b) Boundary alignment of uniform and \MODELNAME with sign transitions estimated by~\cite{twostream}.}
    \label{fig:concept}
\end{figure}

Realizing a sentence as signing motion requires coordinated movements of the body and both hands, rather than merely an ordered list of static poses~\cite{soke,nsa}. These articulators evolve at different temporal rates, while individual signing events and transitions span highly variable durations. These temporal properties pose two key challenges for SLP: \textbf{(1)} motion representation must accommodate articulator-specific temporal dynamics while preserving fine-grained hand motion, and \textbf{(2)} motion generation must account for the variable durations of signing events and transitions.


The first challenge concerns motion representation. Recent approaches encode continuous sign motion as discrete latent tokens for conditional generation~\cite{t2sgpt,signavatars,soke}. However, fine-grained hand articulation and body motion exhibit distinct temporal dependencies~\cite{sandler2012phonological,ma2024multi}. With a shared token rate, coarse tokenization can discard hand detail, whereas fine tokenization increases the token budget for all articulators. This motivates allocating temporal resolution separately to the body and hands.

The second challenge concerns motion generation: determining \emph{how much motion to generate} at each step. Figure~\ref{fig:concept}(a) compares generation strategies by their temporal units and scheduling schemes. Autoregressive models extend motion token by token~\cite{t2sgpt,soke}, whereas full-sequence discrete diffusion denoises tokens in parallel over a predefined temporal horizon~\cite{madis}. Block diffusion combines causal progression across blocks with parallel denoising within each block~\cite{blockdiffusion,signpr}. However, fixed-duration blocks impose uniform generation intervals that may split coherent articulations or merge distinct motion phases. This motivates adaptive block durations that follow the temporal structure of signing~\cite{sandler2010prosody,loos2022simultaneous}. The two challenges are complementary: articulation-specific token rates do not determine generation intervals, and adapting block durations does not address detail lost during tokenization.

We introduce \textbf{\textsc{Blanc}} (\underline{\textbf{BL}}ock-wise \underline{\textbf{A}}daptive diffusio\underline{\textbf{N}} with Articulatory \underline{\textbf{C}}lock), a gloss-free SLP framework that couples multi-rate motion representation with adaptive block diffusion. To address the first challenge, we propose an articulation-decoupled multi-rate vector-quantized motion tokenizer (MRVQ) that encodes the body and both hands as separate discrete streams at articulation-specific temporal resolutions. This preserves fine-grained hand motion while keeping the body representation compact.

To address the second challenge, we extend Block Denoising Diffusion Discrete Language Models (BD3-LMs)~\cite{blockdiffusion} with variable-duration generation blocks. Our generator progresses causally across blocks and denoises tokens in parallel within each block, preserving the distinct temporal rates of the body and hand streams. At inference, generation intervals must be determined without access to the target motion. Our Latent Articulatory Clock (LAC) causally predicts block durations and sequence termination from the source text and generated motion history, enabling generation without a prescribed total sequence length. LAC learns temporal scheduling from motion-derived partitions produced by a deterministic multi-scale teacher. The teacher detects persistent changes in frozen body and hand codebook embeddings, providing temporal supervision without gloss-boundary annotations. Figure~\ref{fig:concept}(b) compares our predicted block boundaries with uniform partitions using CSLR-estimated sign transitions~\cite{twostream} as a reference. Our boundaries align more closely with these estimated transitions, suggesting that motion-informed scheduling better reflects the temporal structure of signing.

To better preserve the linguistic information conveyed by hand articulations, we further propose body-conditioned hand refinement, which remasks and re-predicts selected hand tokens while keeping the initially generated body tokens fixed.

Our contributions are as follows:
\begin{itemize}
    \item We propose MRVQ, an articulation-decoupled multi-rate tokenizer that preserves fine-grained hand motion with compact body representations.
    \item We introduce \textsc{Blanc}, an adaptive block diffusion framework that leverages a motion-supervised LAC to predict variable block durations and sequence termination.
    \item Our framework achieves state-of-the-art performance on PHOENIX-2014T and CSL-Daily, supported by extensive ablation studies.
\end{itemize}
\section{Related Works}

\noindent \textbf{Diffusion-based Sign Language Production.}
Diffusion models~\cite{ddpm,latentdiffusion} have been increasingly adopted for sign language production (SLP), enabling high-fidelity and diverse motion generation.
Early approaches focused on continuous motion modeling, incorporating anatomical graph structures~\cite{nsa}, disentangled bone representations~\cite{signidd}, and joint-aware denoising architectures~\cite{fgdm} to improve motion plausibility.
Recent studies have further explored linguistic conditioning and gloss-free generation to enhance semantic alignment and temporal coherence~\cite{g2p,tdm,signdiff}.
Meanwhile, discrete diffusion approaches have emerged to model sign motion through quantized representations, offering an alternative to continuous motion generation~\cite{signpr,madis,signvip}.

\noindent \textbf{Block-wise Diffusion for Variable-Length Generation.}
Discrete diffusion models enable parallel generation~\cite{sahoo2024mdlm,llada}, while block-wise diffusion combines autoregressive decoding across blocks with parallel denoising within each block~\cite{blockdiffusion}. However, fixed-size blocks impose rigid generation boundaries, motivating recent studies to improve generation efficiency and training-inference alignment~\cite{fastdllm2,d2f,blockwisesft}. Other approaches explore adaptive block scheduling based on token confidence and semantic content~\cite{adablock,swordsman,dcdm,adaptiveblockdiffusion}. Despite these advances, existing approaches primarily focus on text generation, without explicitly considering the temporal dynamics of articulated motion.

\noindent \textbf{Temporal Structure Learning for Sign Language.}
Temporal structure has been studied in sign language understanding through temporal segmentation~\cite{signsegment}, sign boundary detection~\cite{boundary_detect}, and gloss-to-frame alignment for recognizing continuous signing~\cite{iterative_slr}.
In sign language production, such temporal modeling has been further exploited for sequence progression, gloss-wise duration prediction, length regulation, and co-articulation~\cite{progtrans,natea,signingatscale}.
Recently, explicit duration modeling has been used to construct temporally grounded linguistic conditions for generation~\cite{signer}.

\section{Method}
\subsection{Background: Block Diffusion Modeling}
Autoregressive (AR) models generate sequences token by token, requiring $L$ sequential decoding steps for a sequence of length \(L\).
Diffusion language models~\cite{llada, mmada, dream}, in contrast, update multiple token positions jointly but require iterative denoising.
BD3-LMs~\cite{blockdiffusion} combine these paradigms by generating autoregressively across blocks while denoising the tokens within each block in parallel.

Given a token sequence $\mathbf{x}$, BD3-LMs partition it into $B=L/L'$ consecutive blocks of a fixed length $L'$.
Denoting the $b$-th block by $\mathbf{x}^b$ and all preceding blocks by \(\mathbf{x}^{<b}\), the sequence likelihood factorizes as
\begin{equation}
    \small
    \log p_\theta(\mathbf{x})
    =
    \sum_{b=1}^{B}
    \log p_\theta
    \left(
    \mathbf{x}^{b}\mid\mathbf{x}^{<b}
    \right).
\end{equation}
Each block is generated through a discrete diffusion process conditioned on the previously generated clean blocks.
Thus, BD3-LMs reduce autoregressive dependencies from individual tokens to blocks while enabling parallel generation within each block.

\subsection{Motion Representation and Tokenization}
\noindent \textbf{Data Preparation.}
Following prior works~\cite{signavatars, signavox}, we represent sign language sequences using 3D pose parameters. We employ 3D estimators tailored to distinct anatomical regions: specifically, SMPLer-X~\cite{smplerx} for extracting body pose and WiLoR~\cite{wilor} for capturing hand articulations. To ensure high-fidelity pose estimation, we apply an additional optimization step, as detailed in the supplementary material. Ultimately, each processed sign motion is represented as a continuous sequence 
$\mathbf{S} \in \mathbb{R}^{T \times d_s}$, where $T$ denotes the sequence length and $d_s=133$, consistent with the motion representation used in~\cite{soke,madis}. Further details on the pose fitting procedure and quantitative evaluations of the fitted poses, are provided in the Appendix~\ref{sec:appendix_fitting}.
\begin{figure}[t]
    \centering
    \includegraphics[width=\columnwidth]{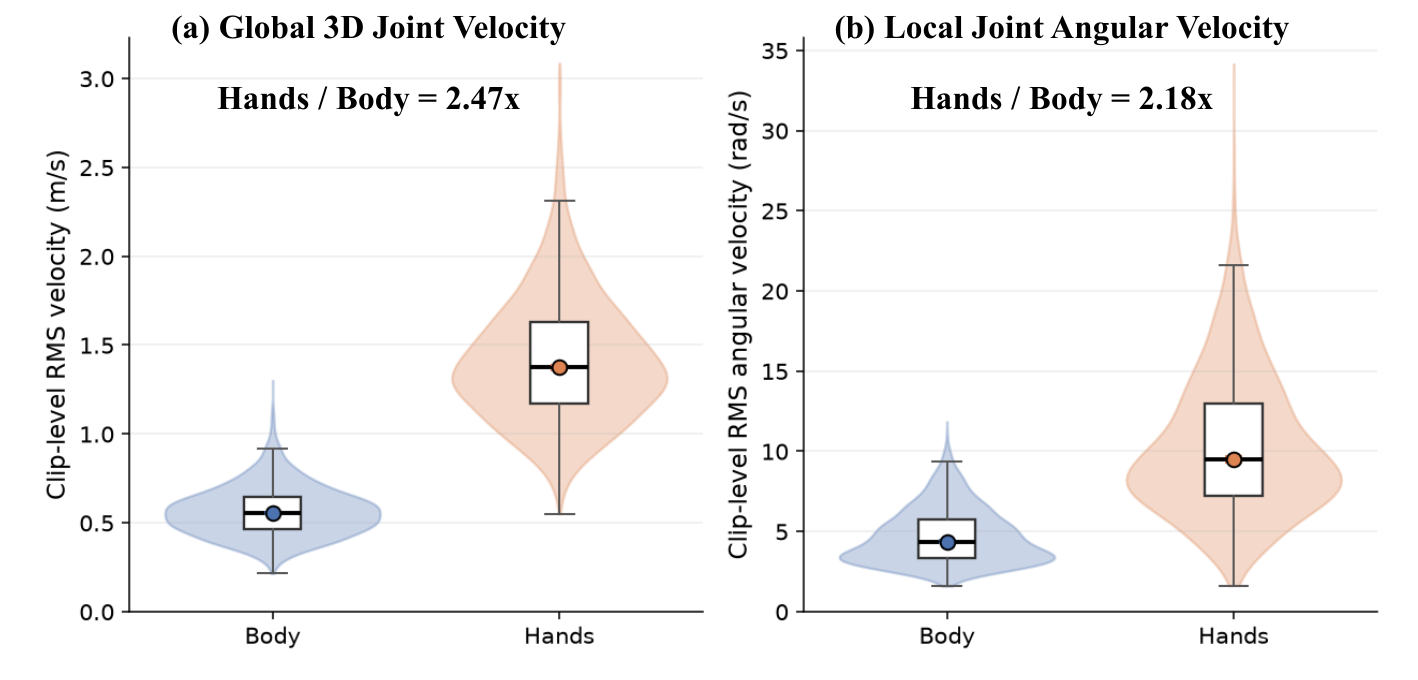}
    \caption{Velocity analysis on PHOENIX-2014T and CSL-Daily. Hands exhibit higher clip-level RMS velocity than the body in both global 3D joint space (a) and local angular space (b), motivating separate temporal rates for body and hand motion.}
    \label{fig:mrvq_analysis}
\end{figure}

\begin{figure*}
    \centering
    \includegraphics[width=\linewidth]{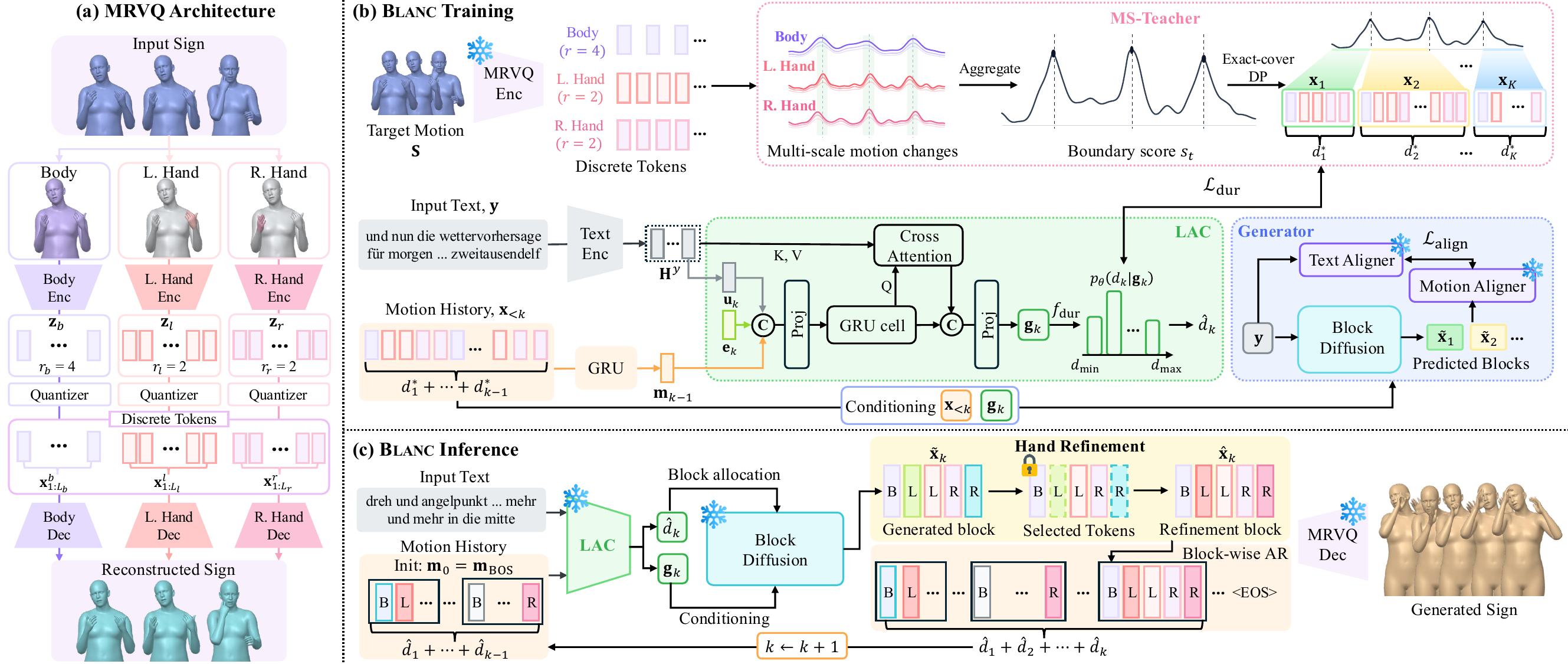}
    \caption{Overview of our method. (a) MRVQ architecture for multi-rate motion tokenization. (b) Training pipeline of \textsc{Blanc} with motion-derived temporal supervision. (c) Inference pipeline with adaptive block generation and body-conditioned hand refinement.}
    \label{fig:model_overview}
\end{figure*}

\noindent \textbf{Articulation-Decoupled Multi-Rate Tokenizer.}
To address the motion representation challenge, we design a discrete representation that accommodates articulator-specific temporal dynamics while preserving fine-grained hand motion.
To this end, we propose a Multi-Rate Vector Quantizer (MRVQ), an articulation-decoupled tokenizer based on VQ-VAE~\cite{vqvae}.
Following prior works~\cite{soke,signavatars}, we model the body, left hand, and right hand with independent VQ-VAEs.
This separation allows us to assign articulator-specific temporal resolutions, as illustrated in Figure~\ref{fig:model_overview}(a).

As shown in Figure~\ref{fig:mrvq_analysis}, hand motion exhibits substantially faster temporal variation than body motion, with $2.47\times$ higher linear velocity and $2.18\times$ higher angular velocity.
Motivated by this observation, we use part-specific temporal compression factors, setting $r_b = 4$ for the body and $r_l = r_r = 2$ for the hands.
Each hand stream therefore has twice the temporal resolution of the body stream.

Given a motion sequence $\mathbf{S}\in\mathbb{R}^{T\times d_s}$, we split it into part-wise streams $\mathbf{S}_p$, where $p\in\{b,l,r\}$.
Each stream is encoded independently as
\begin{equation}
\small
\mathbf{z}_p=\mathcal{E}_p(\mathbf{S}_p)\in\mathbb{R}^{L_p\times d},
\qquad
L_p=\left\lceil T/r_p \right\rceil,
\end{equation}
where $L_p$ denotes the latent sequence length of articulator $p$.
Quantizing these latent vectors with a part-specific codebook of size $K_p$ yields a discrete token sequence
\begin{equation}
\small
\mathbf{x}_p =
\left(x_p^1,\ldots,x_p^{L_p}\right),
\qquad
x_p^i\in\{1,\ldots,K_p\}.
\end{equation}
The three streams jointly form the multi-rate discrete representation
\begin{equation}
\small
\mathbf{x} =
\left(\mathbf{x}_b,\mathbf{x}_l,\mathbf{x}_r\right),
\qquad
N_{\mathrm{tok}} = L_b + L_l + L_r.
\end{equation}
Here, $L_b$, $L_l$, and $L_r$ denote the token lengths of the body, left-hand, and right-hand streams, respectively, while $N_{\mathrm{tok}}$ is the total number of tokens across all three streams.
This representation retains the temporal order within each stream and serves as the discrete input to the subsequent block diffusion model.

The tokenizer is trained with the objective
\begin{equation}
\small
\mathcal{L}_{\mathrm{tok}} =
\sum_{p\in\{b,l,r\}}
\lambda_p\mathcal{L}_{\mathrm{rec}}^{p}
+
\lambda_{\mathrm{fk}}\mathcal{L}_{\mathrm{fk}}
+
\lambda_{\mathrm{emb}}\mathcal{L}_{\mathrm{emb}}.
\end{equation}
Additional details on tokenizer implementation are provided in the Appendix~\ref{app:mrvq_details}.

\subsection{BLANC}
Given the discrete motion representation, we seek to organize generation blocks that better reflect the temporal structure of sign articulation. However, BD3-LMs~\cite{blockdiffusion} employ fixed-size blocks, which may disrupt coherent articulatory patterns by partitioning motion independently of its underlying temporal structure. To address this limitation, we propose \textsc{BLANC}, which leverages motion-derived temporal boundaries to learn adaptive block scheduling. 
The training and inference process of \MODELNAME are illustrated in Figure~\ref{fig:model_overview}(b) and (c), respectively.
Further details can be found Appendix~\ref{app:blanc_details}.

\subsubsection{Learning the Latent Articulatory Clock}
Adapting the duration of each generation block to the temporal structure of signing poses two challenges: identifying suitable generation boundaries without gloss annotations and predicting block durations without access to future motion. Gloss-based segmentation requires linguistic annotations and assumes that lexical boundaries align with suitable motion-generation units. To address these challenges, inspired by~\cite{uboco}, we introduce a Latent Articulatory Clock (LAC) that separates motion-based temporal structure discovery from its causal execution.

\noindent \textbf{Where to split and how long?} The first challenge is to determine \emph{where} a motion sequence should be divided into generation blocks and \emph{how long} each block should persist. We introduce a Multi-scale teacher (MS-Teacher) that derives variable-length block boundaries and duration targets by aggregating motion changes across articulators and temporal scales.

Given a target motion $\mathbf{S}$, we obtain part-wise latent embeddings $\mathbf{q}^{p}$ from MRVQ. At each body-rate timestep $t$, we measure motion variation as the cosine distance between the average latent embeddings before and after $t$, $\bar{\mathbf{q}}^{p,-}_{t,h}$ and $\bar{\mathbf{q}}^{p,+}_{t,h}$, over progressively coarser temporal scales $h\in\{1,2,4\}$.
We aggregate these changes across branches into a motion-change score $\mu_t$ and measure their cross-branch agreement $\rho_t$, yielding the boundary score
\begin{equation}\small
s_t
=
\mu_t
\left(
\frac{1+\rho_t}{2}
\right).
\end{equation}
We select local maxima of $s_t$ as motion-supported boundary candidates.

Rather than treating all candidates as block boundaries, we solve a bounded exact-cover problem using dynamic programming to partition the sequence into $K$ variable-length blocks. The partition covers the entire sequence without gaps or overlaps, constrains each block duration to $[d_{\min}, d_{\max}]$, and favors boundaries with strong motion evidence. Let $\tau_k$ denote the endpoint of the $k$-th block, with $\tau_0=0$ and $\tau_{K}=L_b$. The resulting block duration is
\begin{equation}\small
    d_k^* = \tau_k - \tau_{k-1},
    \qquad
    \sum_{k=1}^{K} d_k^* = L_b,
\end{equation}
which serves as the duration supervision for LAC.

\noindent \textbf{How to execute the clock causally?}
The second challenge is to reproduce this motion-derived temporal structure at inference time, where the target motion and its partition are unavailable. We therefore train a causal LAC controller that predicts the duration of the next block from the input text and previously generated motion.

Given the input text sequence $\mathbf{y}$, a trainable text encoder produces token-level representations $\mathbf{H}^{y}$. 
At generation step $k$, LAC uses accumulated text coverage~\cite{tu2016modeling} to obtain a remaining semantic representation $\mathbf{u}_k$ that emphasizes insufficiently covered content.
In parallel, the motion history $\mathbf{x}_{<k}$ is taken from the target motion up to the teacher-derived boundary during training, while at inference it consists of the previously generated blocks. We map these motion tokens to their frozen MRVQ codebook embeddings and summarize them with a unidirectional GRU~\cite{gru}, obtaining $\mathbf{m}_{k-1}$.

We then combine the remaining semantic representation $\mathbf{u}_k$, $\mathbf{m}_{k-1}$, and a learnable event-position embedding $\mathbf{e}_k$ through $f_{\mathrm{LAC}}$. These representations are projected and used to update a recurrent state with a GRU cell. The updated state then attends to the text representation $\mathbf{H}^{y}$ through cross-attention, and the recurrent state and attended text context are fused to form the latent articulatory event state
\begin{equation}
    \small
    \mathbf{g}_{k}
    =
    f_{\mathrm{LAC}}
    (\mathbf{u}_k,\mathbf{m}_{k-1},\mathbf{e}_k;\mathbf{H}^{y}).
\end{equation}
The event state $\mathbf{g}_k$ captures both the semantic content yet to be expressed and the current motion progression. Based on this state, LAC predicts the duration distribution of the next block:
\begin{equation}\small
    p_{\theta}(d_k | \mathbf{g}_k) = \operatorname{Softmax}(f_{\mathrm{dur}}(\mathbf{g}_k)), 
    \ \
    d_{k} \in \{d_{\min}, \cdots, d_{\max}\}.
\end{equation}
During training, LAC is supervised by motion-derived durations $d_k^*$. At inference, it predicts $\hat{d}_k$ at each block-wise autoregressive step, adapting the generation schedule to the remaining text semantics and previously generated motion.

\begin{table*}[t!]
\centering
\small
\setlength{\tabcolsep}{3pt}
\resizebox{\linewidth}{!}{%
\begin{tabular}{l cc cc cc cc cc cc}
\toprule
\multirow{3}{*}{Method}
& \multicolumn{6}{c}{PHOENIX-2014T}
& \multicolumn{6}{c}{CSL-Daily} \\
\cmidrule(lr){2-7} \cmidrule(lr){8-13}

& \multicolumn{2}{c}{DTW-PA-JPE$\downarrow$}
& \multicolumn{2}{c}{DTW-JPE$\downarrow$}
& \multicolumn{2}{c}{B-T$\uparrow$}
& \multicolumn{2}{c}{DTW-PA-JPE$\downarrow$}
& \multicolumn{2}{c}{DTW-JPE$\downarrow$}
& \multicolumn{2}{c}{B-T$\uparrow$} \\
\cmidrule(lr){2-3}
\cmidrule(lr){4-5}
\cmidrule(lr){6-7}
\cmidrule(lr){8-9}
\cmidrule(lr){10-11}
\cmidrule(lr){12-13}

& Body & Hand
& Body & Hand
& BLEU-4 & ROUGE-L
& Body & Hand
& Body & Hand
& BLEU-4 & ROUGE-L \\
\midrule

T2M$^{\dagger}$~\cite{t2m}
& 36.54 & 3.63
& 35.69 & 24.12
& 0.31 & 5.95
& 28.83 & 3.58
& 30.97 & 18.63
& 0.00 & 10.80 \\

MotionGPT~\cite{motiongpt}
& 8.22 & 1.70
& 8.31 & 9.07
& 5.62 & 16.88
& 10.05 & 2.19
& 9.86 & 10.65
& 0.00 & 9.15 \\

MoMask~\cite{momask}
& 8.20 & 1.58
& 7.91 & 7.69
& 6.44 & 19.81
& 10.58 & 2.15
& 10.51 & 9.61
& 0.57 & 15.89 \\

SOKE (no dict.)~\cite{soke}
& 8.24 & 1.64
& 8.69 & 8.07
& 5.13 & 23.86
& 9.94 & \underline{1.77}
& 9.63 & 8.70
& \textbf{2.66} & \textbf{21.13} \\

SignPR$^{\dagger}$~\cite{signpr}
& 7.87 & 1.87
& 7.67 & 7.43
& 1.04 & 9.82
& 10.63 & 2.37
& 10.75 & 9.72
& 0.27 & 13.06 \\

SignSparK (no KF)~\cite{signspark}
& 8.35 & 1.81
& 8.32 & 7.96
& 1.21 & 10.08
& 9.83 & 2.16
& \textbf{9.17} & 8.78
& 0.32 & 12.82 \\

\rowcolor[HTML]{dbe4ff}
\textsc{Blanc} (\textit{w/o} Refinement)
& \textbf{7.57} & \underline{1.49}
& \textbf{7.31} & \underline{7.01}
& \underline{9.58} & \underline{24.83}
& \textbf{9.67} & \textbf{1.71}
& \underline{9.48} & \textbf{8.19}
& 1.86 & 18.52 \\

\rowcolor[HTML]{dbe4ff}
\textsc{Blanc}
& \underline{7.58} & \textbf{1.48}
& \underline{7.33} & \textbf{7.00}
& \textbf{9.85} & \textbf{25.36}
& \underline{9.75} & \textbf{1.71}
& 9.54 & \underline{8.25}
& \underline{2.13} & \underline{19.05} \\
\specialrule{\heavyrulewidth}{0pt}{\belowrulesep}
\end{tabular}%
}
\caption{
Comparison with gloss-free SLP methods on PHOENIX-2014T and CSL-Daily.
B-T denotes back-translation. \textbf{Bold} and \underline{underlined} indicate the best and second-best results, respectively.
$\dagger$ marks reproduced results.
}
\label{tab:exp_prev_comparison}
\end{table*}
\subsubsection{Block-wise Motion Generation}
\noindent \textbf{Block Diffusion Training.}
During training, each motion-derived interval defines a multi-stream block $\mathbf{x}_k=(\mathbf{x}_{b,k},\mathbf{x}_{l,k},\mathbf{x}_{r,k})$, where $\mathbf{x}_{p,k}$ contains tokens from stream $p$ aligned with the body-rate interval $(\tau_{k-1},\tau_k]$.
We train the block diffusion model to recover randomly masked tokens in $\mathbf{x}_k$, conditioned on the input text $\mathbf{y}$, preceding ground-truth blocks $\mathbf{x}_{<k}$, and the latent articulatory event state $\mathbf{g}_k$.
Bidirectional denoising within each block produces an unrefined prediction $\tilde{\mathbf{x}}_k$.

\noindent \textbf{Training Objective.}
We jointly optimize the block diffusion generator and LAC using the following objective:
\begin{equation}\small
\label{eq:blanc_objective}
    \begin{gathered}
        \mathcal{L}
        =
        \mathcal{L}_{\mathrm{motion}}
        +\lambda_{d}\mathcal{L}_{\mathrm{dur}}
        +\lambda_{\mathrm{len}}\mathcal{L}_{\mathrm{len}}
        +\lambda_{e}\mathcal{L}_{\mathrm{EOS}} \\
        +\lambda_{a}\mathcal{L}_{\mathrm{align}}
        +\lambda_{cf}\mathcal{L}_{\mathrm{cf}}.
    \end{gathered}
\end{equation}
Here, $\mathcal{L}_{\mathrm{motion}}$ supervises masked motion token prediction, while $\mathcal{L}_{\mathrm{dur}}$ trains LAC using motion-derived duration targets $d_k^*$. To preserve semantic consistency, $\mathcal{L}_{\mathrm{align}}$ aligns denoised motion predictions with the input text using a separately pretrained, frozen text--motion aligner. The remaining objectives and implementation details
are provided in the Appendix~\ref{app:blanc_objectives}.

\noindent \textbf{Inference.}
Given the input text $\mathbf{y}$, we initialize the motion-history state as $\mathbf{m}_0=\mathbf{m}_{\mathrm{BOS}}$ for the first block.
At each step $k$, LAC predicts the latent articulatory event state $\mathbf{g}_k$ and duration $\hat{d}_k$ based on the input text and previously generated motion $\hat{\mathbf{x}}_{<k}$.
Conditioned on $\mathbf{y}$, $\hat{\mathbf{x}}_{<k}$, and $\mathbf{g}_k$, block diffusion generates an initial motion block $\tilde{\mathbf{x}}_k$ through a single parallel denoising pass.

To better preserve the linguistic information conveyed by hand articulations, we introduce entropy-guided hand refinement. We select the top $\eta$ fraction of hand tokens within the current block ranked by normalized predictive entropy in descending order. The selected tokens are remasked and re-predicted in a single refinement pass using the same generator, while keeping the body and unselected hand tokens fixed, yielding the refined block $\hat{\mathbf{x}}_k$.
The refined motion updates the history state for subsequent block-wise autoregressive generation until \texttt{EOS} is predicted.
Finally, the frozen MRVQ decoder reconstructs the continuous sign motion.

\begin{figure*}[t]
    \centering
    \includegraphics[width=0.9\linewidth]{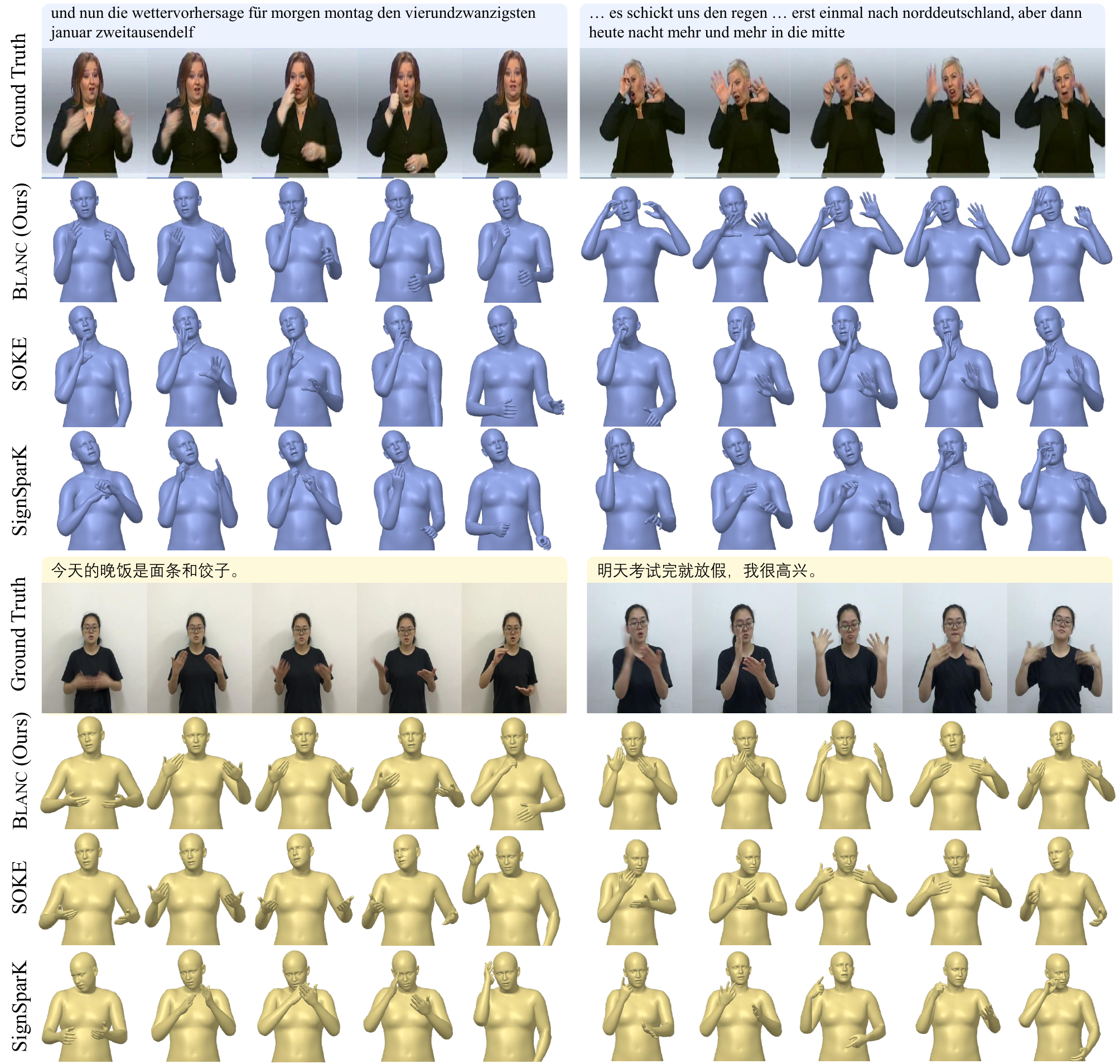}
    \caption{Qualitative comparisons with SOKE and SignSparK on PHOENIX-2014T (top) and CSL-Daily (bottom). Each example shows the input text, reference video frames, and generated signing poses.}
    \label{fig:qualitative_results}
\end{figure*}
\section{Experiments}
\noindent \textbf{Datasets.}
For sign language production, we use two datasets following previous works~\cite{signer, gard}: PHOENIX-2014T (Ph-T)~\cite{phoenix} and CSL-Daily (CSL)~\cite{csl}, which contain approximately 8K and 20K video-text pairs for German and Chinese sign language, respectively. 

\noindent \textbf{Evaluation Metrics.}
To evaluate our model, we assess the generated sign sequences in terms of motion quality and linguistic accuracy. For motion quality, we report Dynamic Time Warping-aligned Joint Position Error (DTW-JPE) and its Procrustes-aligned variant (DTW-PA-JPE). 
For linguistic accuracy, we train a back-translation model~\cite{progtrans} on our 3D sign representations and report BLEU-4~\cite{bleu} and ROUGE-L~\cite{rouge} following~\cite{soke,nsa,signer}. 
We additionally evaluate the reconstruction quality of the tokenizer using Procrustes-Aligned Mean Per Joint Position Error (PA-MPJPE)~\cite{osx,signavatars}.


\subsection{Comparison with State-of-the-Art Methods}
We compare our method with recent gloss-free sign language production approaches on both datasets. Our baselines include general text-to-motion generation methods, such as T2M~\cite{t2m}, MotionGPT~\cite{motiongpt}, and MoMask~\cite{momask}, as well as recent sign-specific approaches, including SOKE~\cite{soke}, SignPR~\cite{signpr}, and SignSparK~\cite{signspark}. For SOKE and SignSparK, we use their gloss-free settings for a fair comparison. 

\begin{table}[t]
\centering
\small
\resizebox{\linewidth}{!}{%
\begin{tabular}{lcccc}
\toprule
\multirow{2}{*}{Method} & \multicolumn{2}{c}{Ph-T (JPE) $\downarrow$} & \multicolumn{2}{c}{CSL (JPE) $\downarrow$} \\
\cmidrule(lr){2-3} \cmidrule(lr){4-5}
 & Body & Hands & Body & Hands \\
 \midrule
 MotionGPT~\cite{motiongpt} & 29.78 & 5.69 & 30.71 & 4.27 \\
 MoMask~\cite{momask} & 23.65 & 5.27 & 25.07 & 4.12 \\
 SOKE~\cite{soke} & 25.63 & 4.66 & 21.56 & 3.41 \\
 \rowcolor[HTML]{dbe4ff}
 Ours (MRVQ) & \textbf{15.48} & \textbf{4.37} & \textbf{11.80} & \textbf{3.06} \\
 \specialrule{\heavyrulewidth}{0pt}{\belowrulesep}
 \end{tabular}%
 }
 \caption{Reconstruction errors of the tokenizer on PHOENIX-2014T (Ph-T) and CSL-Daily (CSL). We report PA-MPJPE to assess reconstruction performance.}
 \label{tab:vqvae_results}
 \end{table}


\noindent \textbf{Quantitative Results for SLP.}
Table~\ref{tab:exp_prev_comparison} compares our method with gloss-free SLP approaches on Ph-T and CSL. General text-to-motion models transfer poorly to sign language, particularly in fine-grained hand articulation and linguistic preservation. \textsc{Blanc}, in contrast, achieves strong motion quality across both datasets, especially for hand motion, while maintaining competitive back-translation performance. Moreover, refinement consistently improves back-translation quality with little change in motion error, indicating that body-conditioned hand refinement enhances linguistic fidelity without sacrificing geometric quality.



\noindent \textbf{Qualitative Results for SLP.}
Figure~\ref{fig:qualitative_results} presents qualitative comparisons with SOKE and SignSparK. Our method more closely matches the ground truth hand positions, particularly in relation to the face and torso. In contrast, SOKE better captures the wrist rotation in some cases, resulting in a hand orientation closer to the reference.


\noindent \textbf{Tokenizer Reconstruction.}
We compare the motion reconstruction performance of our tokenizer against existing methods~\cite{motiongpt, momask, soke} that employ VQ-VAE-based tokenizers, with the results reported in Table~\ref{tab:vqvae_results}. Our proposed tokenizer achieves the lowest reconstruction error across both body and hand regions on two datasets. In particular, it yields substantial improvements in body reconstruction while consistently outperforming prior methods on the hands. These results demonstrate that the articulation-decoupled multi-rate design effectively preserves part-specific motion characteristics and enables more precise motion tokenization.

\subsection{Ablation Studies}
\noindent \textbf{Effect of Adaptive Block Scheduling.}
\begin{table}[t]
\centering
\small
\resizebox{\linewidth}{!}{%
\begin{tabular}{lcccccc}
\toprule
\multirow{3}{*}{\raisebox{-15pt}{\shortstack{Block Size\\($L'$)}}}
& \multicolumn{3}{c}{Ph-T}
& \multicolumn{3}{c}{CSL} \\
\cmidrule(lr){2-4} \cmidrule(lr){5-7}
& \multicolumn{2}{c}{DTW-PA-JPE $\downarrow$}
& \multicolumn{1}{c}{B-T $\uparrow$}
& \multicolumn{2}{c}{DTW-PA-JPE $\downarrow$}
& \multicolumn{1}{c}{B-T $\uparrow$} \\
\cmidrule(lr){2-3} \cmidrule(lr){4-4}
\cmidrule(lr){5-6} \cmidrule(lr){7-7}
& Body & Hands & BLEU-4
& Body & Hands & BLEU-4 \\
\midrule
$L'=1$ (AR) & 9.10 & 1.69 & 6.37 & 11.14 & 2.01 & 0.68 \\
$L'=5$  & 11.33 & 2.10 & 4.37 & 15.17 & 2.59 & 0.98 \\
$L'=10$ & 11.79 & 2.23 & 4.10 & 14.79 & 2.56 & 0.98 \\
$L'=20$ & 10.95 & 2.11 & 4.76 & 14.53 & 2.56 & 1.18 \\
\rowcolor[HTML]{dbe4ff}
\textsc{Blanc}
& \textbf{7.58} & \textbf{1.48} & \textbf{9.85} & \textbf{9.75} & \textbf{1.71} & \textbf{2.13} \\
\specialrule{\heavyrulewidth}{0pt}{\belowrulesep}
\end{tabular}%
}
\caption{Comparison of fixed block sizes and our adaptive block generation. $L'$ denotes the fixed block size, with $L'=1$ corresponding to autoregressive (AR) generation.}
\label{tab:fix_vs_ours}
\end{table}
To assess adaptive block scheduling, we compare \textsc{Blanc} with fixed-block baselines based on BD3-LMs~\cite{blockdiffusion}, using $L'\in\{1,5,10,20\}$.
As shown in Table~\ref{tab:fix_vs_ours}, the fixed-block baselines are sensitive to the choice of block size.
Autoregressive generation ($L'=1$) yields the lowest body and hand errors among the fixed-block settings on both datasets.
On CSL-Daily, however, larger fixed blocks achieve higher BLEU-4 scores despite their higher motion errors.
\textsc{Blanc} achieves the best performance on all reported metrics, improving BLEU-4 from $6.37$ to $9.85$ on PHOENIX-2014T and from $1.18$ to $2.13$ on CSL-Daily over the respective best fixed-block scores.
These results support adapting generation durations to the temporal structure of signing to improve both motion fidelity and linguistic accuracy.


\begin{table}[t]
\centering
\small
\resizebox{\linewidth}{!}{%
\begin{tabular}{lccccc}
\toprule
\multirow{2}{*}{\shortstack{Temporal\\Supervision}}
& \multicolumn{2}{c}{DTW-PA-JPE $\downarrow$}
& B-T $\uparrow$
& \multicolumn{2}{c}{Boundary $\uparrow$} \\
\cmidrule(lr){2-3}
\cmidrule(lr){4-4}
\cmidrule(lr){5-6}
& Body & Hands & BLEU-4 & Precision & F1 \\
\specialrule{\lightrulewidth}{\aboverulesep}{0pt}
\rowcolor{gray!10}
\multicolumn{6}{l}{\textbf{\textit{PHOENIX-2014T}}} \\
Uniform
& 7.93 & 1.51 & 8.98 & 55.49 & 37.89 \\ 
Shuffled
& 7.63 & 1.50 & 8.54 & 54.57 & 37.26 \\
Single-Scale
& \textbf{7.58} & \textbf{1.48} & 8.44 & 60.14 & 41.07 \\
\rowcolor[HTML]{dbe4ff}
Multi-Scale (Ours)
& \textbf{7.58} & \textbf{1.48} & \textbf{9.85} & \textbf{62.06 }& \textbf{42.37} \\

\specialrule{\heavyrulewidth}{0pt}{0pt}
\rowcolor{gray!10}
\multicolumn{6}{l}{\textbf{\textit{CSL-Daily}}} \\
Uniform
& 10.10 & 1.81 & 1.06 & 47.60 & 29.91 \\
Shuffled
& 9.77 & 1.75 & 1.91 & 47.43 & 29.81 \\
Single-Scale
& \textbf{9.63} & 1.72 & 1.93 & 59.41 & 37.34 \\
\rowcolor[HTML]{dbe4ff}
Multi-Scale (Ours)
& 9.75 & \textbf{1.71} & \textbf{2.13} & \textbf{60.28} & \textbf{37.88} \\

\specialrule{\heavyrulewidth}{0pt}{\belowrulesep}
\end{tabular}%
}
\caption{Ablation of temporal supervision strategies. Boundary precision. and F1 evaluate alignment with CSLR-derived boundaries.}
\label{tab:temp_super}
\end{table}
\noindent \textbf{Effect of Motion-Derived Temporal Supervision.}
Table~\ref{tab:temp_super} compares three LAC supervision variants: \emph{Uniform} uses equal-duration blocks, \emph{Shuffled} randomly reorders motion-derived durations, and \emph{Single-Scale} extracts boundaries using $\mathcal{H}=\{1\}$ instead of $\{1,2,4\}$.
We additionally evaluate boundary alignment using precision and F1 against CSLR-estimated boundaries~\cite{twostream}, with further details provided in the Appendix~\ref{sec:evaluation_metrics}.

Our multi-scale supervision achieves the highest BLEU-4 and boundary alignment scores on both datasets while maintaining competitive motion errors.
Compared with \emph{Single-Scale}, the improvements in both semantic fidelity and boundary alignment suggest that multi-scale temporal cues provide more informative supervision beyond pose accuracy alone.
\newcommand{\up}{\rlap{\textsubscript{\textcolor{red}{$\blacktriangle$}}}}
\newcommand{\down}{\rlap{\textsubscript{\textcolor{blue}{$\blacktriangledown$}}}}
\newcommand{\eq}{\rlap{\textsubscript{\textcolor{gray}{\rule{0.6em}{0.6ex}}}}}

\begin{table}[t]
\centering
\small
\resizebox{\linewidth}{!}{%
\begin{tabular}{lcccccc}
\toprule
\multirow{3}{*}{\raisebox{-15pt}{\shortstack{Setting}}}
& \multicolumn{3}{c}{Ph-T}
& \multicolumn{3}{c}{CSL} \\
\cmidrule(lr){2-4} \cmidrule(lr){5-7}

& \multicolumn{2}{c}{DTW-PA-JPE $\downarrow$}
& B-T $\uparrow$
& \multicolumn{2}{c}{DTW-PA-JPE $\downarrow$}
& B-T $\uparrow$ \\
\cmidrule(lr){2-3} \cmidrule(lr){4-4}
\cmidrule(lr){5-6} \cmidrule(lr){7-7}

& Body & Hands & BLEU-4
& Body & Hands & BLEU-4 \\
\specialrule{\lightrulewidth}{\aboverulesep}{0.1pt}
\rowcolor{gray!10}
\multicolumn{7}{l}{\textbf{\textit{LAC Component Ablation}}} \\
\addlinespace[1.5pt]
\textit{w/o} Motion Feedback
& 7.81 & 1.54 & 9.64
& 9.87 & 1.91 & 0.23 \\

\textit{w/o} Alignment Loss
& 7.61 & 1.49 & 9.54
& 9.79 & 1.72 & 1.54 \\

\specialrule{\lightrulewidth}{\aboverulesep}{0.1pt}
\rowcolor{gray!10}
\multicolumn{7}{l}{\textbf{\textit{Refinement Method}}} \\
\addlinespace[1.5pt]

\textit{w/o} Refinement
& 7.57 & 1.49 & 9.58
& 9.67 & 1.71 & 1.86 \\ \cdashline{1-7}

\rule{0pt}{2.4ex}Body + Hands
& 7.59\down
& 1.49\eq
& 9.19\down
& 9.78\down
& \textbf{1.70}\up
& 1.82\down \\

Body Refinement
& 7.57\eq
& 1.48\up
& 9.08\down
& 9.85\down
& 1.72\down
& 1.76\down \\

Hands Refinement (Ours)
& \textbf{7.58}\down
& \textbf{1.48}\up
& \textbf{9.85}\up
& \textbf{9.75}\down
& 1.71\eq
& \textbf{2.13}\up \\
\bottomrule
\end{tabular}%
}
\caption{
Ablation of LAC components and refinement strategies.
For refinement, \textcolor{red}{$\blacktriangle$}, \textcolor{blue}{$\blacktriangledown$}, and \raisebox{0.3ex}{\textcolor{gray}{\rule{0.6em}{0.6ex}}} denote improvement, degradation, and no change relative to \textit{w/o} refinement, respectively.
}
\label{tab:ablation}
\end{table}

\noindent \textbf{Effect of LAC Components and Refinement.}
Table~\ref{tab:ablation} examines motion feedback ($\mathbf{x}_{<k}$), semantic alignment ($\mathcal{L}_{\mathrm{align}}$), and inference-time refinement. Removing motion feedback increases motion errors on both datasets and substantially degrades back-translation performance on CSL, underscoring the importance of conditioning the clock on motion history. Removing the alignment loss also lowers BLEU-4 on both datasets, supporting the role of semantic supervision in preserving linguistic content.

Hand-only refinement improves BLEU-4 on both datasets, with slightly higher body DTW-PA-JPE but comparable hand accuracy. In contrast, body-only and joint body--hand refinement reduce BLEU-4, suggesting that hand-focused refinement improves linguistic fidelity with minimal geometric trade-offs.

\section{Conclusion}
We presented \MODELNAME, a gloss-free sign language production framework that combines multi-rate motion tokenization with adaptive block diffusion. MRVQ represents body and hand motion at distinct temporal resolutions, while the Latent Articulatory Clock learns from motion-derived temporal supervision to predict block durations and sequence termination. Body-conditioned hand refinement further improves hand articulation using the generated body motion as context. Experiments on PHOENIX-2014T and CSL-Daily demonstrate strong motion fidelity and competitive back-translation performance, supporting the effectiveness of combining articulation-specific temporal resolutions with adaptive generation intervals.

\section*{Limitations}
Our method relies on reconstructed 3D sign motion due to the limited availability of 3D annotations in existing sign language datasets. Consequently, reconstruction errors, particularly in fine-grained hand articulations, may affect generation quality. Moreover, while back-translation provides an automatic measure of linguistic fidelity, it may not fully capture the naturalness and comprehensibility of generated signing. Our evaluation does not include human assessments by fluent signers, which remains an important direction for future work. Finally, although our method does not require gloss supervision, our boundary evaluation relies on CSLR-derived alignments, which may not precisely reflect underlying articulatory transitions.

\bibliography{custom}

\clearpage
\appendix



\startcontents[appendix]
\thispagestyle{empty}

\begin{center}
    {\large\bfseries Appendix Contents}\\[0.5em]
    {\normalsize BLANC: Adaptive Temporal Scheduling}
\end{center}

\vspace{1em}

\begingroup
\hypersetup{linkcolor=black}

\titlecontents{section}
    [0em]
    {\small\bfseries\addvspace{0.65em}}
    {\contentslabel{1.5em}}
    {}
    {\hfill\contentspage}

\titlecontents{subsection}
    [1.5em]
    {\footnotesize\addvspace{0.2em}}
    {\contentslabel{2.5em}}
    {}
    {\hfill\contentspage}

\printcontents[appendix]{}{1}{}

\endgroup

\newpage

\section{3D Pose Fitting and Refinement}
\label{sec:appendix_fitting}
Following prior works~\cite{signactor, soke}, we curate frame-level SMPL-X~\cite{smplx} annotations from monocular RGB videos using state-of-the-art methods for 3D body and hand reconstruction. Given a monocular RGB video, we first estimate a rough SMPL-X pose with SMPLer-X~\cite{smplerx} and substitute its hand poses with high-fidelity MANO~\cite{mano} predictions from WiLoR~\cite{wilor}. Since WiLoR provides only a right-hand MANO model and estimates left hands on horizontally flipped inputs, we map the left-hand rotations back to the SMPL-X left-hand frame using the mirror transform $M=\mathrm{diag}(-1,1,1)$, \textit{i.e.}, $R\mapsto MRM$, and recover each wrist's local rotation by inverting the body kinematic chain.

We then refine the initial SMPL-X pose by optimizing the upper-body joint rotations and a sequence-level weak-perspective camera, while keeping the global orientation, lower body, shape, jaw, expression, hands, and wrists fixed. The optimization minimizes the reprojection error between MediaPipe 2D keypoints~\cite{mediapipe} and projected SMPL-X joints for the shoulders, elbows, wrists, and nose, excluding lower-body landmarks. The reprojection loss is compactly written as
\begin{equation}
\small
  \mathcal{L}_{\mathrm{rec}}
  =
  \left\|
  J^{2D} - \pi(\hat{J}^{3D})
  \right\|_{1,\rho,W},
\end{equation}
where $\hat{J}^{3D}$ denotes the predicted SMPL-X joints, $J^{2D}$ denotes the corresponding MediaPipe keypoints, and $\pi(J)=sJ_{xy}+\mathbf{t}$ is the weak-perspective projection. The notation $\|\cdot\|_{1,\rho,W}$ denotes a visibility-weighted robust $L_1$ loss with the Geman--McClure penalty $\rho_\sigma(x)=\frac{\sigma^2x^2}{\sigma^2+x^2}$.

Since frame-wise reprojection fitting can produce temporally unstable motion due to noisy 2D detections and depth ambiguity, we additionally impose a temporal smoothness loss on the reconstructed SMPL-X vertices and joints:
\begin{equation}\small
  \mathcal{L}_{\mathrm{temp}}
  =
  \frac{1}{F-1}
  \sum_{f=2}^{F}
  \left(
  \|V_f - V_{f-1}\|_1
  +
  \|J_f - J_{f-1}\|_1
  \right),
\end{equation}
where $V_f$ and $J_f$ denote the SMPL-X vertices and joints at frame $f$, respectively.

We also use an $L_2$ pose regularization term to suppress implausible upper-body poses:
\begin{equation}\small
  \mathcal{L}_{\mathrm{reg}}
  =
  \frac{1}{|\mathcal{U}|}
  \sum_{j\in\mathcal{U}}
  \|\theta_j\|_2^2,
\end{equation}
where $\mathcal{U}$ is the set of optimized upper-body joints.

The final fitting objective is defined as
\begin{equation} \small
  \mathcal{L}
  =
  \lambda_{\mathrm{rec}}\mathcal{L}_{\mathrm{rec}}
  +
  \lambda_{\mathrm{temp}}\mathcal{L}_{\mathrm{temp}}
  +
  \lambda_{\mathrm{reg}}\mathcal{L}_{\mathrm{reg}}.
\end{equation}
We minimize this objective with Adam and keep the substituted hands fixed throughout the optimization.

\begin{figure}[t]
    \centering
    \includegraphics[width=\linewidth]{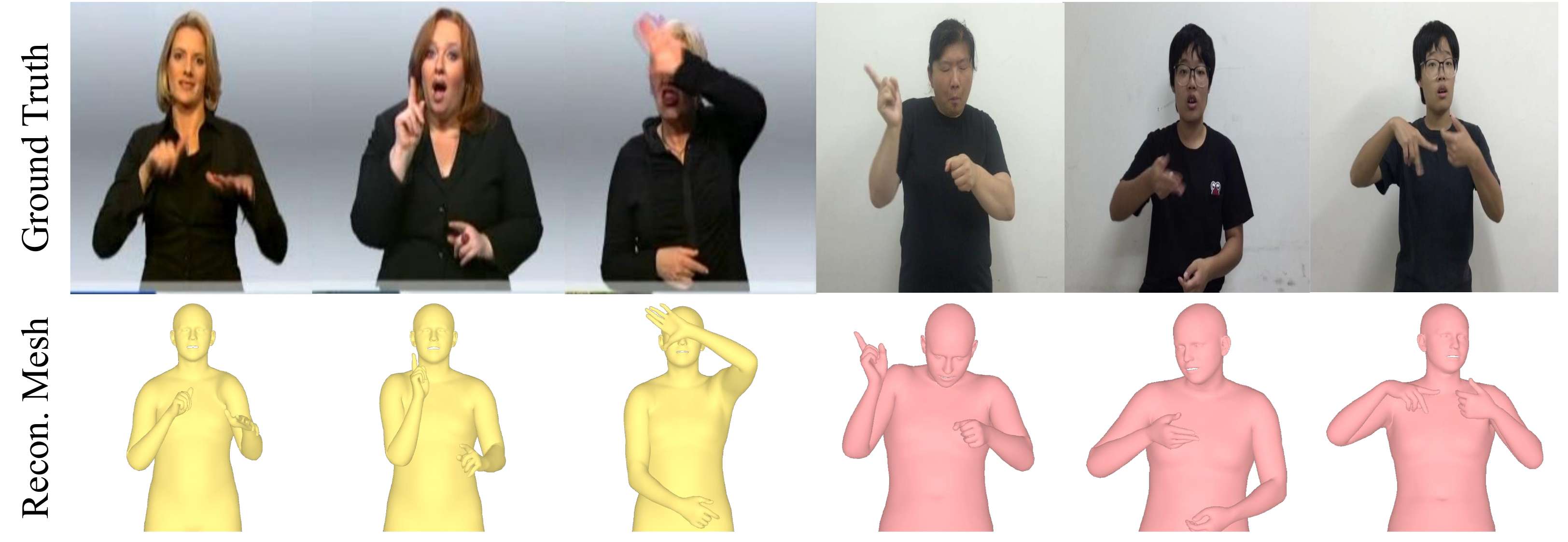}
    \caption{Qualitative results between ground truth video frames and reconstructed meshes obtained from the proposed our fitting pipeline on  Phoenix-2014T (yellow) and CSL-Daily (red).}
    \label{fig:3d_fitting_vis}
\end{figure}

Following~\cite{soke}, we quantitatively evaluate our fitting procedure on the SGNify dataset~\cite{sgnify}, with results reported in Table~\ref{tab:3d_fitting}. Our method achieves strong reconstruction performance, particularly for the hands, obtaining the lowest error on the left hand and competitive performance on the right hand. Meanwhile, it maintains comparable accuracy for the body against strong 3D pose estimation baselines. These results demonstrate that our fitting procedure effectively preserves fine-grained hand articulations while maintaining accurate body pose reconstruction. Qualitative results in Figure~\ref{fig:3d_fitting_vis} further demonstrate that our method faithfully captures both body pose and fine-grained hand articulations.
\begin{table}[t]
\centering
\small
\resizebox{\linewidth}{!}{%
\begin{tabular}{l|ccc}
\toprule
\textbf{Method} & \textbf{Body}$\downarrow$ & \textbf{Left Hand}$\downarrow$ & \textbf{Right Hand}$\downarrow$ \\
\midrule
FrankMoCap~\cite{frankmocap} & 78.07 & 20.47 & 19.62 \\
PIXIE~\cite{pixie}           & 60.11 & 25.02 & 22.42 \\
PyMAF-X~\cite{pymafx}        & 68.61 & 21.46 & 19.19 \\
SMPLify-X~\cite{smplx}    & 56.07 & 22.23 & 18.83 \\
SGNify~\cite{sgnify}         & 55.63 & 19.22 & 17.50 \\
OSX~\cite{osx}               & 47.32 & 18.34 & 18.12 \\
NSA~\cite{nsa}               & \underline{46.42} & 16.17 & 15.23 \\
SMPLer-X~\cite{smplerx}      & \textbf{42.98} & 28.28 & 24.86 \\
SOKE~\cite{soke}             & 46.73 & \underline{10.55} & \textbf{8.94} \\
\rowcolor[HTML]{dbe4ff}
Ours                          & 46.48 & \textbf{9.17} & \underline{9.83} \\
\specialrule{\heavyrulewidth}{0pt}{\belowrulesep}
\end{tabular}%
}
\caption{Reconstruction errors on SGNify mocap dataset~\cite{sgnify}. We report mean per vertex errors in mm.}
\label{tab:3d_fitting}
\end{table}
\section{Additional Details of the Multi-Rate Vector Quantizer}
\label{app:mrvq_details}

Following the notation in the main paper and Figure~\ref{fig:model_overview}(a), we detail the codebook assignment, motion reconstruction, and individual training losses of MRVQ.

\subsection{Quantization and Codebook Optimization}
\label{app:mrvq_quantization}

Each latent feature is quantized using a part-specific codebook $\mathcal{C}_p=\{\mathbf{c}_p^i\}_{i=1}^{N_p}$.
We use $N_b=96$ codebook entries for the body and $N_l=N_r=192$ entries for the hand streams.
Before quantization, the latent vectors and codebook embeddings are $\ell_2$-normalized; thus, nearest-neighbor assignment is equivalent to cosine-similarity-based quantization:
\begin{equation}
\small
    x_p^t
    =
    \mathop{\arg\min}_{j\in\{1,\ldots,N_p\}}
    \left\|
    \frac{\mathbf{z}_p^t}{\|\mathbf{z}_p^t\|_2}
    -
    \frac{\mathbf{c}_p^j}{\|\mathbf{c}_p^j\|_2}
    \right\|_2,
    \qquad
    \hat{\mathbf{z}}_p^t=\mathbf{c}_p^{x_p^t}.
\end{equation}
This produces an articulator-specific discrete sequence $\mathbf{x}_p=(x_p^1,\ldots,x_p^{L_p})$.

Following VQ-VAE~\cite{vqvae2}, we update the codebook embeddings using exponential moving averages (EMA) rather than direct gradient-based optimization.
We additionally monitor code utilization and reinitialize under-utilized entries during training to prevent codebook collapse and improve token diversity.

\subsection{Motion Reconstruction}
\label{app:mrvq_reconstruction}
The interleaved tokens defined in the main paper are first separated into their part-wise streams and mapped back to codebook embeddings. The resulting latent sequences are decoded by the part-specific decoders $\mathcal{D}_p$.
Each decoder upsamples its latent sequence to the original frame rate, after which the reconstructed streams are concatenated to recover the complete motion:
\begin{equation}
\small
    \hat{\mathbf{S}}_p
    =
    \mathcal{D}_p(\hat{\mathbf{z}}_p),
    \qquad
    \hat{\mathbf{S}}
    =
    \left[
    \hat{\mathbf{S}}_b,
    \hat{\mathbf{S}}_l,
    \hat{\mathbf{S}}_r
    \right].
\end{equation}

\subsection{Training Objectives}
\label{app:mrvq_objectives}

The tokenizer objective in Equation~\ref{eq:mrvq_objective} combines reconstruction, forward-kinematics, and commitment losses.
We define each component below; optimization settings and loss weights are reported in Appendix~\ref{sec:hyperparameter_settings} and Table~\ref{tab:mrvq_hyperparameters}.

\noindent\textbf{Reconstruction loss.}
For each articulator, the reconstruction term is a Smooth-$\ell_1$ loss between the reconstructed and ground-truth motion representations:
\begin{equation}
\small
    \mathcal{L}_{\mathrm{rec}}^{p}
    =
    \frac{1}{T}
    \sum_{t=1}^{T}
    \operatorname{SmoothL1}
    \left(
    \hat{\mathbf{S}}_p^{t},
    \mathbf{S}_p^{t}
    \right).
\end{equation}

\noindent\textbf{Forward-kinematics loss.}
To preserve the articulated 3D geometry, we additionally apply a forward-kinematics loss.
Let $\operatorname{FK}(\cdot)$ denote the SMPL-X~\cite{smplx} forward-kinematics operation that maps motion parameters to 3D joint positions.
We minimize the Euclidean distance between the joints obtained from the reconstructed and ground-truth motions:
\begin{equation}
\small
    \mathcal{L}_{\mathrm{fk}}
    =
    \frac{1}{TJ}
    \sum_{t=1}^{T}
    \sum_{j=1}^{J}
    \left\|
    \operatorname{FK}(\hat{\mathbf{S}})_{t,j}
    -
    \operatorname{FK}(\mathbf{S})_{t,j}
    \right\|_2.
\end{equation}

\noindent\textbf{Commitment loss.}
We further use the standard VQ commitment loss~\cite{vqvae} to keep the encoder outputs close to their assigned codebook embeddings:
\begin{equation}
\small
    \mathcal{L}_{\mathrm{emb}}
    =
    \sum_{p\in\{b,l,r\}}
    \left\|
    \mathbf{z}_p
    -
    \operatorname{sg}[\hat{\mathbf{z}}_p]
    \right\|_2^2,
\end{equation}
where $\operatorname{sg}[\cdot]$ denotes the stop-gradient operation.

\section{Additional Details of \MODELNAME}
\label{app:blanc_details}

This section expands the training and inference procedures in Figure~\ref{fig:model_overview}(b) and (c), following the notation of the main paper.
The motion sequence $\mathbf{x}=(\mathbf{x}^{1},\ldots,\mathbf{x}^{L_b})$ contains $L_b$ body-rate units, each comprising one body token and two tokens from each hand.
A block $\mathbf{x}_k$ spans units $\tau_{k-1}+1$ through $\tau_k$ and contains $5d_k$ tokens, where $d_k=\tau_k-\tau_{k-1}$.
MRVQ is frozen throughout generator training.

\subsection{Multi-Scale Motion Teacher}
\label{app:blanc_teacher}

\noindent\textbf{Multi-scale boundary evidence.}
The MS-Teacher identifies temporal changes in the frozen MRVQ embedding space without gloss annotations.
To compare the three articulators at a common temporal resolution, we retain the body embedding and average the two hand embeddings within each body-rate unit:
\begin{equation}
\small
\begin{gathered}
\small
    \mathbf{v}_b^t = \operatorname{norm}(\mathbf{c}_b^{x_b^t}),\\
    \small
    \mathbf{v}_p^t = \operatorname{norm}\!\left(
    \frac{\mathbf{c}_p^{x_p^{2t-1}}+\mathbf{c}_p^{x_p^{2t}}}{2}
    \right),\quad p\in\{l,r\},
\end{gathered}
\end{equation}
where $\mathbf{c}_p^i$ is a codebook embedding and $\operatorname{norm}$ denotes $\ell_2$ normalization.
We use both the original sequence and a normalized three-unit moving average, providing complementary sensitivity to abrupt and persistent changes.

For each articulator, view, and radius $h\in\{1,2,4\}$, we compute the cosine distance between the mean embeddings on either side of a candidate cut $t$.
Writing these means as $\bar{\mathbf{z}}^{p,-}_{t,h}$ and $\bar{\mathbf{z}}^{p,+}_{t,h}$, as in the main paper, the evidence for branch $j$ is
\begin{equation}
\small
    a_{t,j}=1-\cos\!\left(
    \bar{\mathbf{z}}^{p,-}_{t,h},
    \bar{\mathbf{z}}^{p,+}_{t,h}\right).
\end{equation}
The left and right neighborhoods contain at most $h$ units and are truncated at the sequence endpoints.
We normalize each branch across the interior cuts using its median and median absolute deviation (MAD):
\begin{equation}
\small
    b_{t,j}=\sigma\!\left(
    \operatorname{clip}\!\left(
    \frac{a_{t,j}-\operatorname{med}_{t'}a_{t',j}}
    {\kappa_{\mathrm{MAD}}\,\operatorname{MAD}_{t'}(a_{t',j})},-6,6
    \right)\right),
\end{equation}
where $\kappa_{\mathrm{MAD}}$ is the Gaussian consistency factor for MAD, with the MAD lower-bounded by $10^{-4}$.
This normalization makes boundary evidence comparable across articulators and scales.
Let $w_j$ denote normalized branch weights, with total weights $0.35$, $0.325$, and $0.325$ assigned to the body, left hand, and right hand, respectively, and equal weights across views and radii within each part.
We obtain
\begin{equation}
\small
\begin{gathered}
    \mu_t = \sum_j w_j b_{t,j},\\
    \rho_t = \exp\!\left(
    -\frac{\sqrt{\sum_j w_j(b_{t,j}-\mu_t)^2}}{0.25}
    \right),\\
    s_t = \mu_t\frac{1+\rho_t}{2}.
\end{gathered}
\label{eq:blanc_teacher_score}
\end{equation}
Thus, a boundary receives a high score when motion change is both pronounced and consistent across branches.

\noindent\textbf{Bounded exact-cover partition.}
The candidate set $\mathcal{C}$ contains local maxima of $s_t$ above a sequence-specific score quantile.
We then select the partition by maximizing
\begin{equation}
\small
\begin{aligned}
    \max_{K,\tau_1,\ldots,\tau_{K-1}}\;&
    \sum_{k=1}^{K-1} R(\tau_k),\\
    \text{subject to }\;&\tau_0=0,\quad \tau_K=L_b,\\
    &d_{\min}\leq\tau_k-\tau_{k-1}\leq d_{\max},
\end{aligned}
\label{eq:blanc_partition}
\end{equation}
where
\begin{equation}
\small
    R(t)=s_t-\beta-\gamma\,\mathbb{I}[t\notin\mathcal{C}].
\end{equation}
The boundary penalty $\beta$ discourages excessive segmentation, while $\gamma$ penalizes cuts unsupported by the candidate set without excluding cuts needed to satisfy the duration bound.
The number of blocks $K$ is determined jointly with their boundaries.
Dynamic programming solves this problem using
\begin{equation}
\small
    V(t)=\max_{\substack{d_{\min}\leq d\leq d_{\max}\\t-d\geq0}}
    \bigl[V(t-d)+R(t-d)\bigr],
\end{equation}
with $V(0)=0$, $R(0)=0$, and unreachable states assigned $-\infty$.
Backtracking from $L_b$ yields the duration targets $d_k^*$.
We use $\gamma=0.35$ for both datasets; $(\beta,\text{quantile})$ is $(0.50,0.55)$ for PHOENIX-2014T and $(0.40,0.45)$ for CSL-Daily.
Duration bounds are reported in Table~\ref{tab:blanc_hyperparameters}.

We also assign each duration a confidence $c_k$ from its ending boundary.
For $k<K$, this is $s_{\tau_k}\rho_{\tau_k}$, multiplied by $0.25$ for non-candidate cuts and clipped to $[0.05,1]$; the final block has $c_K=1$.
These confidences reduce the influence of weak temporal targets during duration supervision.
The teacher is deterministic and is used only during training.

\subsection{Latent Articulatory Clock}
\label{app:blanc_clock}

\noindent\textbf{Remaining semantics and motion history.}
Let $\mathbf{H}^{y}=(\mathbf{H}^{y}_1,\ldots,\mathbf{H}^{y}_J)$ be the encoded input text and $v_{k-1,j}$ its accumulated coverage before block $k$, initialized to zero.
The remaining semantic representation in Figure~\ref{fig:model_overview}(b) is
\begin{equation}
\small
\begin{aligned}
    r_{k,j} &= \max(1-v_{k-1,j},0),\\
    \mathbf{u}_k &=
    \frac{\sum_{j=1}^{J}r_{k,j}\mathbf{H}^{y}_j}
    {\max(\sum_{j=1}^{J}r_{k,j},\epsilon)}.
\end{aligned}
\end{equation}
Coverage is a learned measure of semantic progress rather than an explicit word-to-motion alignment.

To summarize motion, we project the five frozen codebook embeddings in each $\mathbf{x}^{t}$ into a shared space and average them to form a window feature $\boldsymbol{\psi}^{t}$.
A unidirectional GRU processes these features in temporal order; its state at the end of block $k-1$ gives $\mathbf{m}_{k-1}$.
The first block uses the learned state $\mathbf{m}_0=\mathbf{m}_{\mathrm{BOS}}$.
Training uses preceding ground-truth blocks, whereas inference uses completed generated blocks.

\noindent\textbf{Event state and duration.}
Using the event-position embedding $\mathbf{e}_k$, LAC updates its recurrent state and attends to the text:
{\small
\begin{equation}
\begin{gathered}
    \mathbf{h}_k = \operatorname{GRUCell}\bigl(\mathbf{P}[\mathbf{u}_k ; \mathbf{m}_{k-1} ; \mathbf{e}_k],\, \mathbf{h}_{k-1}\bigr), \\
    (\mathbf{t}_k, \boldsymbol{\alpha}_k) = \operatorname{CrossAttn}(\mathbf{h}_k, \mathbf{H}^y), \\
    \mathbf{g}_k = \mathbf{F}[\mathbf{h}_k ; \mathbf{t}_k].
\end{gathered}
\end{equation}
}
Here, $P$ and $F$ are learned projections, $[\cdot;\cdot]$ denotes concatenation, and $\boldsymbol{\alpha}_k$ is the attention distribution averaged across heads.
A learned gate $\zeta_k=\sigma(G[\mathbf{g}_k;\mathbf{t}_k])$ updates coverage as
\begin{equation}
\small
    v_{k,j}=v_{k-1,j}+\zeta_k\alpha_{k,j}.
\end{equation}

The duration head uses a bounded discrete hazard parameterization.
Let $\nu_{k,d}\in(0,1)$ be the predicted probability of ending the block at duration $d$, conditional on reaching that duration.
Then
\begin{equation}
\small
\begin{gathered}
    p_\theta(d_k=d\mid\mathbf{g}_k)=
    \\
    \begin{cases}
    \nu_{k,d}\displaystyle\prod_{r=d_{\min}}^{d-1}(1-\nu_{k,r}),
        &d<d_{\max},\\
    \displaystyle\prod_{r=d_{\min}}^{d_{\max}-1}(1-\nu_{k,r}),
        &d=d_{\max}.
    \end{cases}
\end{gathered}
\label{eq:blanc_duration_hazard}
\end{equation}
The hazards are sigmoid outputs of a projection of $\mathbf{g}_k$.
The resulting log masses form $f_{\mathrm{dur}}(\mathbf{g}_k)$ in the main paper's softmax expression.
This construction assigns all probability mass to the permitted duration range.

\noindent\textbf{Sequence termination.}
Termination is evaluated after completing a block, so the decision can depend on the newly committed motion.
Let $\mathbf{u}_{k+1}$ be the remaining semantic summary computed from updated coverage $v_k$, and let $\bar r_{k+1}=J^{-1}\sum_j\max(1-v_{k,j},0)$.
The stopping probability is
\begin{equation}
\small
    q_k=\sigma\!\left(f_{\mathrm{EOS}}[
    \mathbf{g}_k;\mathbf{m}_k;\mathbf{u}_{k+1};\bar r_{k+1}]
    \right).
\label{eq:blanc_eos_probability}
\end{equation}
This separates the duration of the current block from the decision to end the entire sequence.

\subsection{Training Objectives}
\label{app:blanc_objectives}

We expand the objective in Equation~\ref{eq:blanc_objective} below.
Expectations are over training examples and sampled corruptions unless otherwise specified.
We write $\ell_{\mathrm{H}}$ for Smooth-$\ell_1$ loss, averaged over its indicated entries.

\noindent\textbf{Motion generation.}
We use $\mathcal{L}_{\mathrm{motion}}$ as a collective term for masked motion prediction and auxiliary reconstruction and representation regularization.
For each teacher-defined block, the generator recovers masked motion tokens conditioned on the input text $\mathbf{y}$, clean preceding blocks $\mathbf{x}_{<k}$, and the event state $\mathbf{g}_k$.
This objective follows the block diffusion formulation~\cite{blockdiffusion}, preserving causal dependence across blocks while allowing bidirectional interactions within each block.

\noindent\textbf{Duration, length, and termination.}
The confidence-weighted duration objective is
\begin{equation}
\small
    \mathcal{L}_{\mathrm{dur}}=
    -\frac{\sum_{n,k}c_{n,k}
    \log p_\theta(d_{n,k}^*\mid\mathbf{g}_{n,k})}
    {\max(1,\sum_{n,k}c_{n,k})},
\end{equation}
where $n$ indexes examples and $k$ indexes their valid blocks.
To prevent local duration biases from accumulating, we also supervise the expected total duration:
\begin{equation}
\small
\begin{gathered}
    \bar d_k=\sum_{d=d_{\min}}^{d_{\max}}
    d\,p_\theta(d\mid\mathbf{g}_k),\\
    \mathcal{L}_{\mathrm{len}}=
    \mathbb{E}\!\left[\ell_{\mathrm{H}}\!\left(
    \frac{\sum_{k=1}^{K}\bar d_k-L_b}{K},0\right)\right].
\end{gathered}
\end{equation}
Finally, the EOS head is supervised to continue after every non-final block and stop after the final block:
\begin{equation}
\small
    \mathcal{L}_{\mathrm{EOS}}=
    \mathbb{E}\!\left[
    -\sum_{k=1}^{K-1}\log(1-q_k)-\log q_K
    \right].
\end{equation}

\noindent\textbf{Semantic alignment.}
We use a symmetric contrastive objective for paired text and motion embeddings.
For normalized pairs $\{(\mathbf{a}_n,\mathbf{b}_n)\}_{n=1}^{B}$, define
\begin{equation}
\small
\begin{aligned}
    \mathcal{I}_{\tau}(\mathbf{A},\mathbf{B})
    =-\frac{1}{2B}\sum_{n=1}^{B}\Bigg[&
    \log\frac{e^{\mathbf{a}_n^\top\mathbf{b}_n/\tau}}
    {\sum_m e^{\mathbf{a}_n^\top\mathbf{b}_m/\tau}}\\
    +&\log\frac{e^{\mathbf{b}_n^\top\mathbf{a}_n/\tau}}
    {\sum_m e^{\mathbf{b}_n^\top\mathbf{a}_m/\tau}}
    \Bigg].
\end{aligned}
\label{eq:blanc_contrastive}
\end{equation}
The text--motion aligner in Figure~\ref{fig:model_overview}(b) is pretrained on paired text and target MRVQ embeddings and then frozen.
Let $A_y$ and $A_x$ denote its text and motion encoders.
For generator supervision, we construct $\mathbf{Z}^{\mathrm{mix}}$ using predicted normalized latents at masked positions and target codebook embeddings elsewhere, and apply
\begin{equation}
\small
    \mathcal{L}_{\mathrm{align}}=
    \mathcal{I}_{0.07}\!\left(
    A_y(\mathbf{y}),A_x(\mathbf{Z}^{\mathrm{mix}})\right).
\end{equation}
Gradients pass through the frozen motion encoder to the generator's continuous predictions, avoiding non-differentiable token sampling.
Examples with dropped text conditioning are excluded from this loss.

\noindent\textbf{Counterfactual event conditioning.}
To encourage the generator to use text-dependent event states, we form counterfactual states $\mathbf{g}_k^{-}$ from mismatched text within the minibatch while retaining the same motion history and teacher partition.
Let $\ell^{+}$ and $\ell^{-}$ denote the mean masked-motion negative log-likelihoods under the matched and counterfactual event states, respectively.
We optimize
\begin{equation}
\small
    \mathcal{L}_{\mathrm{cf}}=
    \max(0,\delta+\ell^{+}-\ell^{-}),
    \qquad\delta=0.25.
\end{equation}
These likelihoods exclude termination symbols and use unweighted token averages.
The generator's direct text input is kept fixed, so the comparison specifically tests its dependence on the text-conditioned event representation.

\noindent\textbf{Loss weights.}
Table~\ref{tab:blanc_loss_weights} reports the maximum coefficients for the duration, length, EOS, alignment, and counterfactual terms in Equation~\ref{eq:blanc_objective}.
Auxiliary weights are introduced with linear ramps early in training; all active coefficients reach their listed values within the first 2,500 optimization steps.
Optimization and architectural settings are given in Appendix~\ref{sec:hyperparameter_settings}.

\begin{table}[t]
\centering
\small
\setlength{\tabcolsep}{5pt}
\begin{tabular}{@{}ccc@{}}
\toprule
Loss coefficient & Ph-T & CSL \\
\midrule
$\lambda_d$ & 0.20 & 0.25 \\
$\lambda_{\mathrm{len}}$ & 0.25 & 0.75 \\
$\lambda_e$ & 0.10 & 0.10 \\
$\lambda_a$ & 0.10 & 0.10 \\
$\lambda_{cf}$ & 0.05 & 0.05 \\
\bottomrule
\end{tabular}
\caption{Maximum training-loss weights for \MODELNAME on PHOENIX-2014T (Ph-T) and CSL-Daily (CSL). The aggregate motion objective has unit weight in Equation~\ref{eq:blanc_objective}.}
\label{tab:blanc_loss_weights}
\end{table}

\subsection{Inference and Hand Refinement}
\label{app:blanc_inference}

Figure~\ref{fig:model_overview}(c) illustrates inference from text alone.
Starting from $\mathbf{m}_0=\mathbf{m}_{\mathrm{BOS}}$ and zero text coverage, LAC predicts $\mathbf{g}_k$ and selects the most probable duration,
\begin{equation}
\small
    \hat d_k=\mathop{\arg\max}_{d\in\{d_{\min},\ldots,d_{\max}\}}
    p_\theta(d\mid\mathbf{g}_k).
\end{equation}
We allocate $5\hat d_k$ masked positions and jointly predict all body and hand tokens in one parallel denoising pass, obtaining the draft block $\tilde{\mathbf{x}}_k$.
The predicted duration determines the block size; no target sequence length or teacher partition is provided.

\noindent\textbf{Guided sampling.}
Let $\boldsymbol{\ell}$ be the fully conditioned logits, $\boldsymbol{\ell}^{-y}$ the logits with direct text conditioning dropped, and $\boldsymbol{\ell}^{-g}$ the logits with event conditioning dropped.
We use
\begin{equation}
\small
\begin{aligned}
    \boldsymbol{\ell}^{\mathrm{guide}}
    ={}&\boldsymbol{\ell}^{-y}
    +w_y(\boldsymbol{\ell}-\boldsymbol{\ell}^{-y})\\
    &+(w_g-1)(\boldsymbol{\ell}-\boldsymbol{\ell}^{-g}),
\end{aligned}
\end{equation}
with $w_y=2.0$ and $w_g=1.5$.
Tokens are sampled with temperature $1.0$ and nucleus threshold $0.9$.

\noindent\textbf{Entropy-guided hand refinement.}
For a hand-token position $i$ belonging to part $p$, let $P_i$ be the draft distribution obtained from the guided logits over its $N_p$ codebook entries.
Its normalized predictive entropy is
\begin{equation}
\small
    H_i=-\frac{1}{\log N_p}
    \sum_{a=1}^{N_p}P_i(a)\log P_i(a).
\end{equation}
We pool the $4\hat d_k$ hand positions across both hands and select the $\lceil\eta\,4\hat d_k\rceil$ largest entropies, forming $\mathcal{M}_k^{\mathrm{hand}}$.
These positions are remasked and predicted once more using the same generator:
\begin{equation}
\small
\begin{aligned}
    \hat{\mathbf{x}}_{k,\mathcal{M}_k^{\mathrm{hand}}}
    \sim p_\theta\bigl(&\mathbf{x}_{k,\mathcal{M}_k^{\mathrm{hand}}}\mid
    \tilde{\mathbf{x}}_{k,\overline{\mathcal{M}}_k^{\mathrm{hand}}},\\
    &\mathbf{y},\hat{\mathbf{x}}_{<k},\mathbf{g}_k,\hat d_k\bigr).
\end{aligned}
\end{equation}
All body tokens and unselected hand tokens remain fixed, allowing uncertain hand articulations to be revised in the context of the draft body trajectory.
We use $\eta=1.0$ on PHOENIX-2014T and $\eta=0.15$ on CSL-Daily.
Refinement introduces neither additional parameters nor a separate training objective.

The refined block $\hat{\mathbf{x}}_k$ updates the motion-history GRU before the EOS decision in Equation~\ref{eq:blanc_eos_probability}.
Generation stops when $q_k\geq0.65$.
As finite decoding safeguards, we allow at most 32 blocks and restrict duration choices to those that fit within the remaining sequence capacity.
Generation also stops if no permitted duration fits.
Finally, the complete generated token sequence is decoded by the frozen MRVQ decoder to obtain continuous motion $\hat{\mathbf{S}}$.
Neither the motion teacher nor the semantic aligner is used at inference.

\section{Experimental Details}
\subsection{Dataset Documentation and Statistics}
\label{sec:dataset_documentation}
PHOENIX-2014T~\cite{phoenix} contains German Sign Language weather forecasts paired with German text, while CSL-Daily~\cite{csl} contains Chinese Sign Language sentences covering everyday topics, paired with Chinese text. Both datasets provide continuous signing sequences and gloss annotations. Our reconstructed 3D poses are derived from these videos and retain the source datasets' language, domain, and signer coverage without expanding their demographic coverage.

The standard training, validation, and test splits contain approximately 7.1K, 0.5K, and 0.6K video--text pairs for PHOENIX-2014T, and 18.4K, 1.1K, and 1.2K pairs for CSL-Daily, respectively.

\subsection{Evaluation Metrics}
\label{sec:evaluation_metrics}

\noindent\textbf{DTW-JPE.}
We evaluate motion in 3D joint space, allowing generated and reference sequences to have different lengths.
For a part with $J$ joints, let $\tilde{\mathbf{P}}_{i,j}$ and $\tilde{\mathbf{Q}}_{u,j}$ denote generated and reference joint positions at frames $i$ and $u$, respectively, after pelvis centering for the body or wrist centering for each hand.
The frame-pair cost is the mean per-joint position error:
\begin{equation}\small
    C_{i,u}=\frac{1}{J}\sum_{j=1}^{J}
    \left\|\tilde{\mathbf{P}}_{i,j}-\tilde{\mathbf{Q}}_{u,j}\right\|_2.
\end{equation}
Following the cumulative DTW convention used in our comparisons with SOKE~\cite{soke}, we report
\begin{equation}\small
    \operatorname{DTW\text{-}JPE}
    =\min_{\pi\in\Pi}\sum_{(i,u)\in\pi}C_{i,u},
\end{equation}
where $\Pi$ contains monotonic warping paths connecting the endpoints of the two sequences.
This score accumulates local MPJPE along the optimal path without dividing by the path length.

\noindent\textbf{DTW-PA-JPE.}
This variant additionally removes differences in global scale, rotation, and translation through similarity Procrustes alignment.
Let $\mathcal{T}_{i,u}$ be the least-squares similarity transform aligning generated frame $i$ to reference frame $u$.
Using the original, uncentered joint positions $\mathbf{P}$ and $\mathbf{Q}$, we compute

\begin{equation}
\small
\begin{gathered}
    C^{\mathrm{PA}}_{i,u} = \frac{1}{J}\sum_{j=1}^{J} \left\|\mathcal{T}_{i,u}(\mathbf{P}_{i,j})-\mathbf{Q}_{u,j}\right\|_2, \\
    \operatorname{DTW\text{-}PA\text{-}JPE} = \min_{\pi\in\Pi}\sum_{(i,u)\in\pi}C^{\mathrm{PA}}_{i,u}.
\end{gathered}
\end{equation}

For both metrics, body error is evaluated on 12 upper-body/head joints and hand error on 21 points per hand.
For body Procrustes alignment, the transform is fitted to the full SMPL-X joint set before selecting the evaluation joints; each hand is aligned independently.
We average the two hand scores and report the mean over test sequences, with lower values indicating better motion fidelity.
For tokenizer reconstruction, PA-MPJPE instead averages Procrustes-aligned joint errors at corresponding frames, without temporal warping.

\noindent\textbf{BLEU-4 and ROUGE-L.}
We assess linguistic preservation by translating generated motion back into text using the same pretrained back-translation model for all compared methods.
BLEU-4~\cite{bleu} measures corpus-level $n$-gram agreement up to order four, while ROUGE-L~\cite{rouge} measures longest-common-subsequence overlap with the reference text.
We use the SacreBLEU-based BLEU and ROUGE-L implementations distributed with SOKE,\footnote{\url{https://github.com/2000ZRL/SOKE/tree/main/mGPT/metrics}} with word-level evaluation for PHOENIX-2014T and character-level evaluation for CSL-Daily.
Both scores are reported on a 0--100 scale, with higher values indicating better performance.

\noindent\textbf{Boundary precision and F1.}
We compare block boundaries with pseudo-reference boundaries derived from CSLR-estimated gloss spans~\cite{twostream}, placing each reference boundary midway between consecutive spans.
Excluding sequence endpoints, we perform one-to-one matching within a tolerance of $\pm4$ motion frames.
Precision measures the fraction of proposed boundaries that are matched, and F1 balances precision with reference-boundary recall.
Counts are pooled across test sequences before computing the scores.
These metrics assess agreement with recognition-derived temporal structure; the pseudo-references are used only for evaluation.

\subsection{Baseline Implementation Details}
We retrain MotionGPT, MoMask, SOKE, and SignSparK~\cite{motiongpt, momask, soke, signspark} from scratch using their official implementations and recommended training configurations. For methods without directly usable implementations, we reproduce them based on the architectural and training details in the original papers.

We implement \MODELNAME based on the official BD3-LMs codebase~\cite{blockdiffusion}\footnote{\url{https://github.com/kuleshov-group/bd3lms}}, adapting it for text-conditioned sign language production.

\subsection{Hardware, Model Size, and Computational Budget}
\label{sec:model_size_compute}
Our computing infrastructure consists of four NVIDIA RTX A6000 GPUs.

MRVQ contains approximately 59.0M trainable parameters. The \MODELNAME training model contains approximately 295M parameters in total, of which 246.2M are trainable. These counts exclude MRVQ; the total includes the frozen 42.2M-parameter semantic aligner and other auxiliary modules. Parameter counts are nearly identical across the two datasets.

For each dataset, training \MODELNAME on four NVIDIA RTX A6000 GPUs takes approximately 12 hours, corresponding to 48 GPU-hours. Thus, the two main generator training runs require approximately 96 GPU-hours in total. These estimates exclude tokenizer training, semantic-aligner pretraining, preprocessing, evaluation, baseline training, hyperparameter search, and ablation experiments.

\subsection{Hyperparameter Settings}
\label{sec:hyperparameter_settings}
Tables~\ref{tab:mrvq_hyperparameters} and~\ref{tab:blanc_hyperparameters} summarize the main hyperparameters for MRVQ and \MODELNAME, respectively. We train a separate tokenizer and generator for each dataset and freeze MRVQ during generator training.

\noindent\textbf{MRVQ.}
We train MRVQ on 64-frame motion windows using Smooth-$\ell_1$ reconstruction loss. The FK loss weight is linearly increased to 10 over the first 20 epochs. Codebooks are updated with EMA, and the same hyperparameters are used for both datasets.

\noindent\textbf{\MODELNAME.}
The generator is trained for 10,000 optimization steps with linear learning-rate warm-up followed by cosine decay to 10\% of the peak learning rate. The evaluated checkpoints are taken at 9,000 steps for PHOENIX-2014T and 10,000 steps for CSL-Daily. At inference, each block is generated in parallel, followed by one body-conditioned hand refinement pass. We remask all hand tokens on PHOENIX-2014T and the most uncertain 15\% on CSL-Daily, keeping body tokens fixed within the refinement pass.
We report generation results from a fixed trained checkpoint using sampling seed 42.

\begin{table}[t]
\centering
\small
\setlength{\tabcolsep}{4pt}
\begin{tabular}{@{}lc@{}}
\toprule
Hyperparameter & Value \\
\midrule
Hidden / code dimension & 512 / 512 \\
Transformer layers / heads & 4 / 8 \\
Temporal compression (body / hands) & 4 / 2 \\
Codebook size (body / each hand) & 96 / 192 \\
Training window (frames) & 64 \\
\midrule
Optimizer & AdamW \\
Learning rate & $2\times10^{-4}$ \\
Adam $\beta_1,\beta_2$ & $(0.9,0.99)$ \\
Weight decay & 0 \\
Batch size & 32 \\
Training epochs & 500 \\
LR schedule & Warm-up + cosine \\
Warm-up steps & 1,000 \\
Gradient clipping norm & 1.0 \\
\midrule
Reconstruction weights (body / L / R) & 1 / 2 / 2 \\
FK loss weight & 10 \\
Commitment coefficient & 0.25 \\
Codebook EMA decay & 0.99 \\
\bottomrule
\end{tabular}
\caption{MRVQ hyperparameters, shared across PHOENIX-2014T and CSL-Daily. Transformer settings apply to both the encoder and decoder. L and R denote the left and right hands.}
\label{tab:mrvq_hyperparameters}
\end{table}

\begin{table}[t]
\centering
\small
\setlength{\tabcolsep}{3pt}
\begin{tabular}{@{}lcc@{}}
\toprule
Hyperparameter & Ph-T & CSL \\
\midrule
Generator dimension & \multicolumn{2}{c}{768} \\
Generator layers / heads & \multicolumn{2}{c}{12 / 12} \\
LAC hidden / event dimension & \multicolumn{2}{c}{512 / 256} \\
Dropout & \multicolumn{2}{c}{0.1} \\
Block duration (body-rate units) & 2--11 & 2--18 \\
MS-Teacher radii & \multicolumn{2}{c}{$\{1,2,4\}$} \\
\midrule
Optimizer & \multicolumn{2}{c}{AdamW} \\
Learning rate & \multicolumn{2}{c}{$1\times10^{-4}$} \\
Adam $\beta_1,\beta_2$ & \multicolumn{2}{c}{$(0.9,0.95)$} \\
Weight decay & \multicolumn{2}{c}{0.01} \\
Batch size per device & \multicolumn{2}{c}{16} \\
Gradient accumulation steps & \multicolumn{2}{c}{2} \\
Training steps & \multicolumn{2}{c}{10,000} \\
LR schedule & \multicolumn{2}{c}{Warm-up + cosine} \\
Warm-up steps & 500 & 200 \\
Gradient clipping norm & \multicolumn{2}{c}{1.0} \\
\midrule
Text / event guidance scale & \multicolumn{2}{c}{2.0 / 1.5} \\
Sampling temperature / top-$p$ & \multicolumn{2}{c}{1.0 / 0.9} \\
Hand refinement passes & \multicolumn{2}{c}{1} \\
Hand remasking ratio & 1.0 & 0.15 \\
\bottomrule
\end{tabular}
\caption{\MODELNAME hyperparameters on PHOENIX-2014T (Ph-T) and CSL-Daily (CSL). One body-rate unit corresponds to four motion frames. Refinement is applied only at inference.}
\label{tab:blanc_hyperparameters}
\end{table}

\section{Additional Experiments}
\subsection{Qualitative Results of MRVQ}
\begin{figure*}[t]
    \centering
    \includegraphics[width=\linewidth]{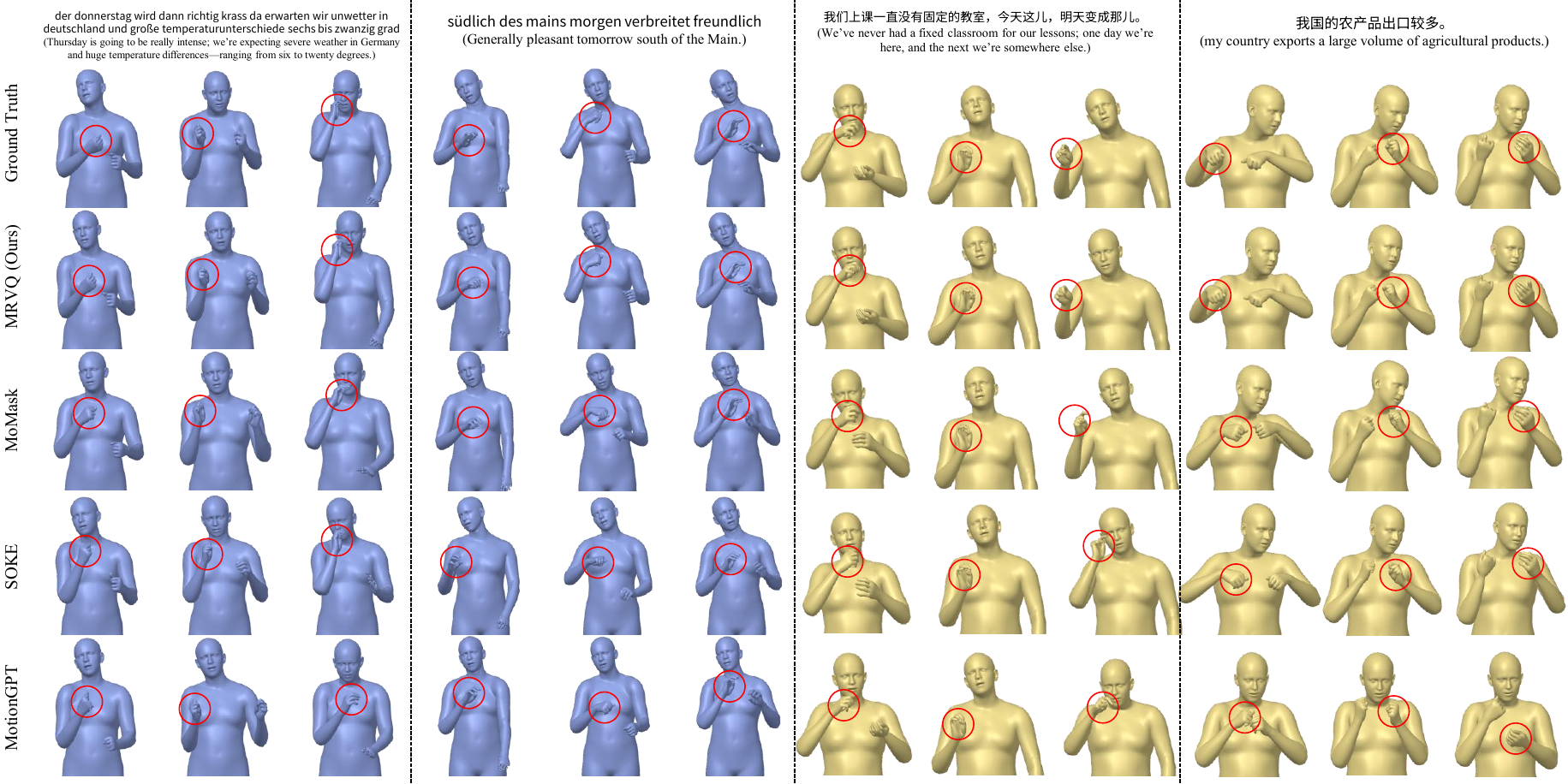}
    \caption{Qualitative comparison of motion reconstruction on PHOENIX-2014T (blue) and CSL-Daily (yellow). Red circles highlight differences in hand positioning and fine-grained articulation across methods.}
    \label{fig:qual_vq_recon}
\end{figure*}
Figure~\ref{fig:qual_vq_recon} compares MRVQ reconstructions with those of MotionGPT, MoMask, and SOKE. In the highlighted examples, MRVQ more closely preserves the reference hand configurations and their positions relative to the face and torso, while maintaining the overall upper-body posture. These visual comparisons complement the quantitative reconstruction results in Table~\ref{tab:vqvae_results}, illustrating the preservation of fine-grained articulatory details through tokenization and reconstruction.

\subsection{Semantic Preservation of Motion Tokenizers}
To evaluate how well each motion tokenizer preserves linguistic information, we tokenize the ground-truth motion using the tokenization schemes adopted by existing methods, reconstruct the motion, and measure its performance with the same back-translator. We additionally report feature-level back-translation performance by directly feeding the ground-truth motion into the back-translator, providing a reference for quantifying the degradation introduced by tokenization. As shown in Table~\ref{tab:back_translation}, our MRVQ consistently achieves the best back-translation performance across both datasets, with substantially smaller degradation from the feature-level results than existing tokenizers. This indicates that the proposed articulation-decoupled multi-rate tokenization better preserves linguistically meaningful information during discretization and reconstruction.

\begin{table}[t]
\centering
\small
\resizebox{\linewidth}{!}{%
\begin{tabular}{lcccc}
\toprule
\multirow{2}{*}{Method}
& \multicolumn{2}{c}{Ph-T$\uparrow$}
& \multicolumn{2}{c}{CSL$\uparrow$} \\
\cmidrule(lr){2-3} \cmidrule(lr){4-5}
& BLEU-4 & ROUGE-L & BLEU-4 & ROUGE-L \\
\specialrule{\lightrulewidth}{\aboverulesep}{0pt}
\rowcolor{gray!15}
\multicolumn{5}{l}{\textbf{\textit{Feature Space}}} \\

GT
& 13.36 & 31.39
& 7.35 & 31.09 \\

\rowcolor{gray!15}
\multicolumn{5}{l}{\textbf{\textit{Token Space}}} \\

MotionGPT
& 9.95 & 25.67 
& 4.97 & 25.81 \\

MoMask
& 10.07 &  26.37
& 5.04 & 26.10 \\

SOKE
& 11.04 & 27.30
& 5.55 & 26.55 \\

\rowcolor[HTML]{dbe4ff}
MRVQ (Ours)
& \textbf{12.49} & \textbf{30.27}
& \textbf{6.21} & \textbf{29.08} \\

\specialrule{\heavyrulewidth}{0pt}{\belowrulesep}
\end{tabular}%
}
\caption{Back-translation performance of motion tokenizers. GT denotes the feature-level reference without tokenization.}
\label{tab:back_translation}
\end{table}

\subsection{Ablation of Refinement}
\label{app:refinement_ablation}
\begin{figure}[t]
    \centering
    \includegraphics[width=\linewidth]{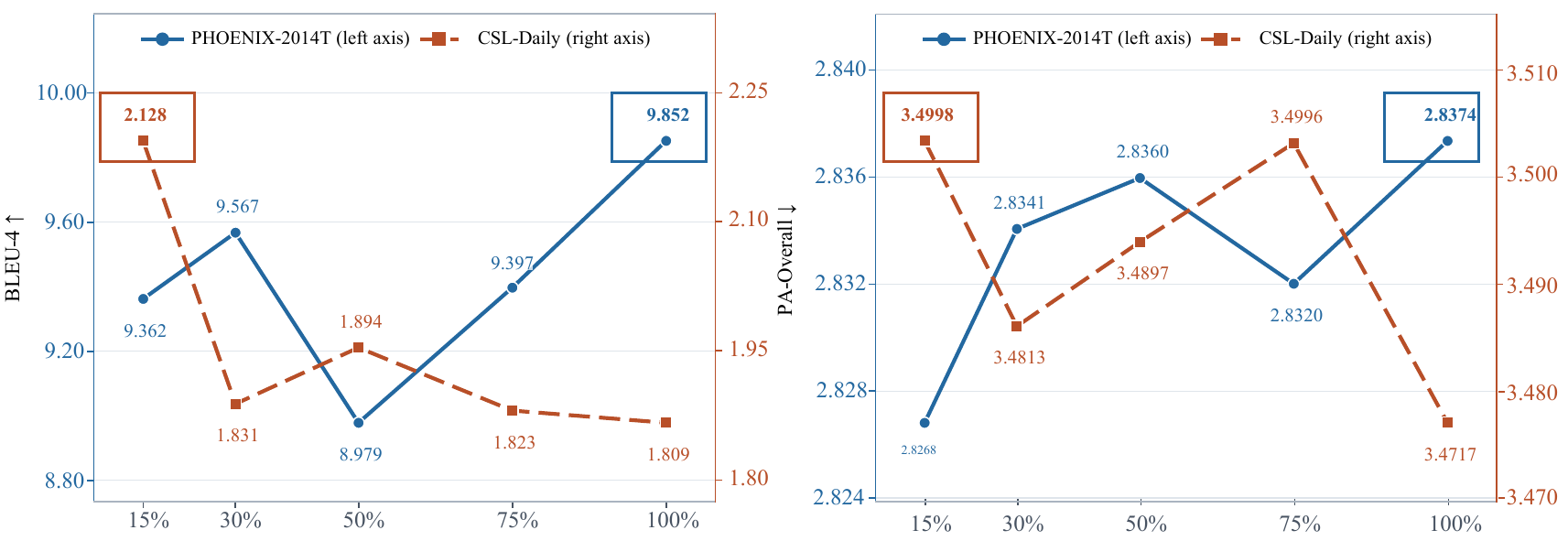}
    \caption{Effect of the hand refinement ratio on BLEU-4 (left; higher is better) and overall DTW-PA-JPE (right; lower is better). Blue curves use the left axes for PHOENIX-2014T, and orange curves use the right axes for CSL-Daily. Boxes mark our selected ratios: 100\% for PHOENIX-2014T and 15\% for CSL-Daily.}
    \label{fig:refinement_ratio}
\end{figure}
Figure~\ref{fig:refinement_ratio} examines the fraction of hand tokens remasked during the single refinement pass, with body tokens kept fixed.
The effect depends on the dataset: BLEU-4 is highest at 100\% refinement on PHOENIX-2014T and 15\% on CSL-Daily, corresponding to our selected configurations.
However, the lowest overall DTW-PA-JPE occurs at 15\% and 100\%, respectively.
Thus, the configurations with the strongest back-translation performance do not minimize reference-based geometric error.
This trade-off suggests that the two metrics capture complementary aspects of generated signing, and that increasing the refinement ratio does not consistently improve both.

\section{Discussion}
\subsection{Broader Impacts and Ethical Considerations}
\MODELNAME may support more accessible communication for Deaf and Hard-of-Hearing communities by advancing text-driven sign language production. However, generated signing may convey incorrect meanings, and limited training data may underrepresent diverse signing styles and language varieties. Practical deployment should therefore involve evaluation with fluent signers and consultation with Deaf communities. Such systems should complement professional interpreters, with their limitations clearly communicated to users. Generated signing should be identified as synthetic to avoid attributing unintended statements to real signers.

\subsection{Data Privacy, Licenses, and Release}
\label{sec:artifact_licenses}
The use and release of sign language data should respect signers' privacy and applicable usage conditions. Our implementation builds on BD3-LMs under Apache-2.0, while the SMPLer-X code used for pose extraction is provided under the non-commercial S-Lab License 1.0. CSL-Daily is governed by a non-commercial research agreement that restricts redistribution. Other third-party datasets and models remain subject to their respective usage and distribution terms.

We plan to release our code, trained checkpoints, and reconstructed 3D poses with applicable license and attribution notices. Release of dataset-derived artifacts is subject to verification of the original terms and any necessary permissions from the rights holders.

\subsection{Use of AI Assistants}
We used ChatGPT and Codex to assist with translation and grammar correction during manuscript preparation. The authors reviewed and revised all AI-assisted content and take full responsibility for the accuracy and integrity of the paper.

\end{document}